\documentclass[final]{cvpr}

\usepackage{times}
\usepackage{epsfig}
\usepackage{graphicx}
\usepackage{amsmath}
\usepackage{amssymb}

\usepackage[utf8]{inputenc} %
\usepackage[T1]{fontenc}    %
\usepackage{url}            %
\usepackage{booktabs}       %
\usepackage{amsfonts}       %
\usepackage{nicefrac}       %
\usepackage{microtype}      %
\usepackage{xcolor}         %
\usepackage{graphicx}
\usepackage{caption}
\usepackage{booktabs}
\usepackage{floatrow}
\usepackage{comment}
\usepackage{wrapfig}
\usepackage{subcaption}
\usepackage{enumerate}
\usepackage{enumitem}
\usepackage{siunitx}

\usepackage[font={footnotesize}]{caption}
\usepackage{subcaption}
\newcommand*{\ptrain}{\ensuremath{P_\mathrm{train}}}
\newcommand*{\ptest}{\ensuremath{P_\mathrm{test}}}

\usepackage[pagebackref=true,breaklinks=true,colorlinks,bookmarks=false]{hyperref}

\usepackage{cleveref}  %

\def\cvprPaperID{8234} %
\def\confYear{CVPR 2021}

\begin{document}

\title{On Robustness and Transferability of Convolutional Neural Networks }

\author{%
  Josip Djolonga\thanks{Shared first authorship. Please send e-mail correspondence to \texttt{\string{josipd,lucic\string}@google.com}.} \quad Jessica Yung\textsuperscript{*} \quad Michael Tschannen\textsuperscript{*} \quad Rob Romijnders \quad Lucas Beyer \\
  Alexander Kolesnikov \quad Joan Puigcerver \quad Matthias Minderer \quad Alexander D'Amour \\ Dan Moldovan \quad Sylvain Gelly \quad Neil Houlsby \quad Xiaohua Zhai \quad Mario Lucic \\
  Google Research, Brain Team
}

\maketitle

\begin{abstract}
Modern deep convolutional networks (CNNs) are often criticized for not generalizing under distributional shifts. However, several recent breakthroughs in transfer learning suggest that these networks can cope with severe distribution shifts and successfully adapt to new tasks from a few training examples. 
In this work we study the interplay between out-of-distribution and transfer performance of modern image classification CNNs for the first time and investigate the impact of the pre-training data size, the model scale, and the data preprocessing pipeline. We find that increasing both the training set and model sizes significantly improve the distributional shift robustness. Furthermore, we show that, perhaps surprisingly, simple changes in the preprocessing such as modifying the image resolution can significantly mitigate robustness issues in some cases. Finally, we outline the shortcomings of existing robustness evaluation datasets and introduce a synthetic dataset \textsc{SI-Score} we use for a systematic analysis across factors of variation common in visual data such as object size and position.
\end{abstract}

\vspace{-5mm}
\section{Introduction}
Deep convolutional networks have attained impressive results across a plethora of visual classification benchmarks \cite{bit,tan2019efficientnet} where the training and testing distributions match. In the real world, however, the conditions in which the models are deployed can often differ significantly from the conditions in which the model was trained. It is thus imperative to understand the impact \emph{dataset shifts} \cite{quionero2009dataset} have on the performance of these models. This problem has gained a lot of traction and several systematic investigations have shown unexpectedly high sensitivity of image classifiers to various dimensions, including photometric perturbations~\cite{hendrycks2018benchmarking}, natural perturbations obtained from video data~\cite{imnetvid}, as well as model-specific adversarial perturbations \cite{goodfellow2014explaining}.

\begin{figure}[t]
    \centering
    \includegraphics[width=\textwidth]{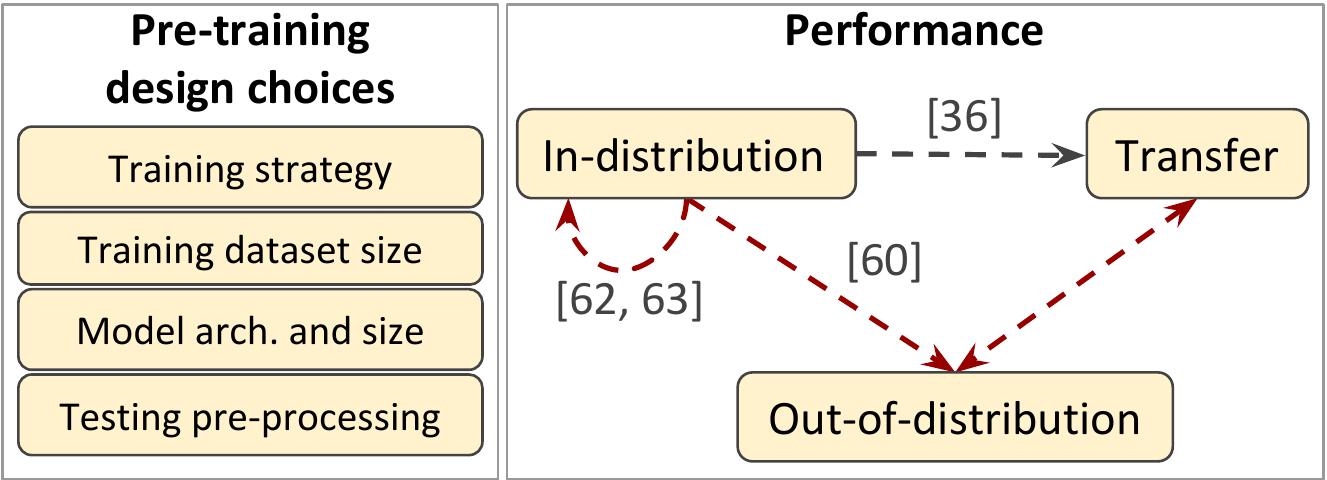}\vspace{-2mm}
    \caption{We explore the fundamental interplay between in-distribution performance, out-of-distribution (OOD) performance, and transfer learning performance (red arrows in the graph on the right), with respect to the major design choices listed on the left. The relationship between in-distribution and OOD performance is highly under-explored along these axes, whereas that between OOD and transfer performance has not been studied before to the best of our knowledge.
\label{fig:big_picture}}
\end{figure}
The problem of dataset shift, or \emph{out-of-distribution (OOD) generalization}, is closely related to a learning paradigm known as \emph{transfer learning} \cite[\S 13]{storkey2009training}. In transfer learning we are interested in constructing models that can improve their performance on some target task by leveraging data from different related problems. In contrast, under dataset shift one assumes that there are two environments, namely training and testing~\cite{storkey2009training}, with the constraint that the model cannot be adapted using data from the target environment. As a consequence, the two environments typically have to be more similar and their differences more structured than in the transfer setting (c.f.\ \Cref{sec:background}).

In the context of transfer learning, detailed scaling laws characterizing the interplay between the in-distribution and transfer performance as a function of pre-training data set size, model size, architectural choices such as normalization, and transfer strategy have been established recently \cite{DBLP:conf/cvpr/KornblithSL19, zhai2019largescale, bit}. Model and dataset scale were identified as key factors for transfer performance. The similarities between transfer learning and OOD generalization suggests that these axes are also relevant for OOD generalization and raises the question of what the corresponding scaling laws are. While some axes have been partially explored by prior work \cite{hendrycks2018benchmarking, noisystudent}, the big picture is largely unknown. Even more importantly, is in-distribution performance enough to characterize OOD performance, or can transfer performance give a more fine-grained characterization of OOD performance of a population of models than in-distribution performance? To the best of our knowledge, this question has not been systematically explored before in the literature.

\vspace{1mm}
\noindent\textbf{Contributions}\quad
We systematically investigate the interplay between the in-distribution accuracy of image classification models on the training distribution, their generalization to OOD data (without adaptation), and their transfer learning performance with adaptation in the low-data regime (see Fig.~\ref{fig:big_picture} for an illustration). Specifically: %
\begin{enumerate}[label=(\roman*),nosep]
    \item We present the first meta-analysis of existing OOD metrics and transfer learning benchmarks across a wide variety of models, ranging from self-supervised to fully supervised models with up to 900M parameters. %
We show that increasing the model and data scale disproportionately improves transfer and OOD performance, while only marginally improving the performance on the \textsc{ImageNet} validation set.
\item Focusing on OOD robustness, we analyze the effects of the training set size, model scale, and the training regime and testing resolution, and find that the effect of scale overshadows all other dimensions. 
\item We introduce a novel dataset for fine-grained OOD analysis to quantify the robustness to object size, object location, and object orientation (rotation angle). We believe that this is a first systematic study to show that the models become less sensitive (and hence more robust) to each of these factors of variation as the dataset size and model size increase.
\end{enumerate}

\section{Background}\label{sec:background}

\noindent\textbf{Robustness of image classification models}\quad
Understanding and correcting for dataset shifts are classical problems in statistics and machine learning, and have as such received substantial attention, see e.g.\ the monograph \cite{quionero2009dataset}.
Formally, let us denote the observed variable by $X$ and the variable we want to predict by $Y$.
A dataset shift occurs when we train on samples from $\ptrain(X,Y)$, but are at test time evaluated under a different distribution $\ptest(X,Y)$.
Storkey~\cite{storkey2009training} discusses and precisely defines different possibilities for how \ptrain~and \ptest~can differ.
We are mostly interested in covariate shifts, i.e., when the conditionals $\ptrain(Y | X)=\ptest(Y | X)$ agree, but the marginals $\ptrain(X)$ and $\ptest(X)$ differ. Most robustness datasets proposed in the literature targeting \textsc{ImageNet} models are such instances---the images $X$ come from a source $\ptest(X)$ different from the original collection process $\ptrain(X)$, but the label semantics do not change.
As a robustness score one typically uses the expected accuracy, i.e., $\ptest(Y = f(X))$, where $f(X)$ is the class predicted by the model.

\vspace{1mm}
\noindent\textbf{Dataset shift types}\quad
\textsc{ImageNet-v2} is a recollected version of the \textsc{ImageNet} validation set~\cite{recht2019imagenet}.
The authors attempted to replicate the data collection process, but found that all models drop significantly in accuracy. Recent work attributes this drop to statistical bias in the data collection \cite{engstrom2020identifying}.
\textsc{ImageNet-C} and \textsc{ImageNet-P} \cite{hendrycks2018benchmarking} are obtained by corrupting the \textsc{ImageNet} validation set with classical corruptions, such as blur, different types of noise and compression, and further cropping the images to $224\times 224$.
These datasets define a total of 15 noise, blur, weather, and digital corruption types, each appearing at 5 severity levels or intensities. \textsc{ObjectNet} \cite{barbu2019objectnet} presents a new test set of images collected directly using crowd-sourcing.
\textsc{ObjectNet} is particular as the objects are captured at unusual poses in cluttered, natural scenes, which can severely degrade recognition performance. Given this clutter, and arguably better suitability as a detection than recognition task~\cite{borji2020reobjectnet}, $Y | X$ might be hard to define and the dataset goes beyond a covariate shift. In contrast, the \textsc{ImageNet-A} dataset \cite{hendrycks2019natural} consists of real-world, unmodified, and naturally occurring examples that are misclassified by ResNet models. Hence in addition to the covariate shift due to the data source, this dataset is not model-agnostic and exhibits a strong selection bias \cite{storkey2009training}.

Attempting to focus on naturally occurring invariances, \cite{imnetvid} annotated two video datasets: \textsc{ImageNet-Vid-Robust} and \textsc{YouTube-BB-Robust}, derived from the \textsc{ImageNet-Vid} \cite{deng2009imagenet} and \textsc{YouTube-BB} \cite{real2017youtube} datasets respectively.
In \cite{imnetvid} the authors propose the \texttt{pm}-$k$ metric---given an anchor 
frame and up to $k$ neighboring frames, a prediction is marked as correct only if the classifier correctly classifies all $2k+1$ frames around and including the anchor.
We present the details of each dataset in Appendix~\ref{app:analysis}.

\noindent\textbf{Transferability of image classification models}\quad
In transfer learning~\cite{pan2009survey}, a model might leverage the data it has seen on a related distribution, $P_\mathrm{pre-train}$, to perform better on a new task \ptrain.
Note that in contrast to the covariate shift setting, the disparity between $P_\mathrm{pre-train}$ and the new task is typically larger, but one is further given samples from \ptrain.
While there exist many approaches on how to transfer to the new task, the most common approach in modern deep learning, which we use, is to (i) train a model on $P_\mathrm{pre-train}$ (using perhaps an auxiliary, self-supervised task \cite{dosovitskiy2015discriminative,gidaris2018unsupervised}), and then (ii) train a model on $\ptrain$ by initializing the model weights from the model trained in the first step. 

Recently, a suite of datasets has been collected to benchmark modern image classification transfer techniques \cite{zhai2019largescale}.
The Visual Task Adaptation Benchmark (VTAB) defines 19 datasets with 1000 labeled samples each, categorized into three groups: %
\emph{natural} (most similar to \textsc{ImageNet}) consists of standard natural classification tasks (e.g.,\ CIFAR);
\emph{specialized} contains medical and satellite images;
and \emph{structured} (least similar to \textsc{ImageNet}) consists mostly of synthetic tasks that require understanding of the geometric layout of scenes.
We compute an overall transfer score as the mean across all 19 datasets, as well as scores for each subgroup of tasks.
We provide details for all of the tasks in \Cref{app:analysis}.

\begin{figure*}[t]
\begin{tabular}{ccc}
    \hspace{-0.4cm}
    \includegraphics[height=3.5cm,trim={2.5mm 0 4.5mm 0},clip]{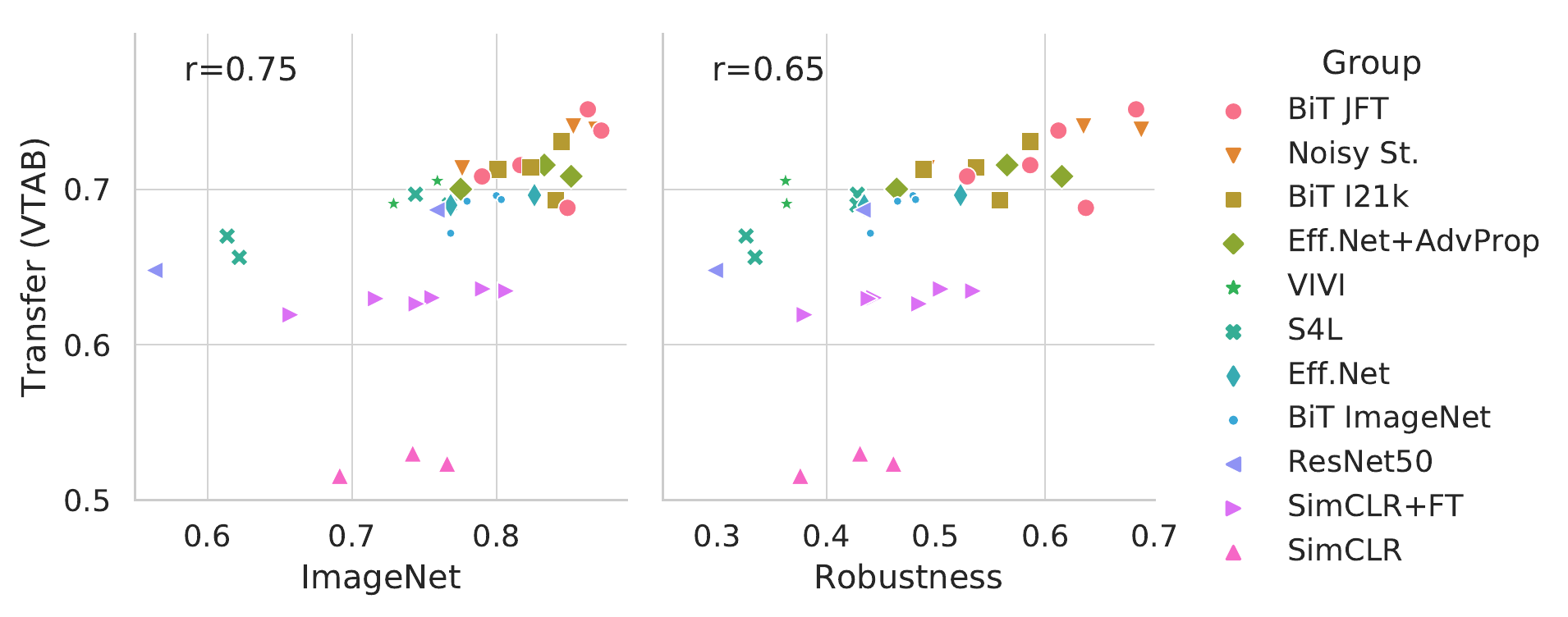}&
    \includegraphics[width=0.4\textwidth,trim={2.5mm 0 0mm 0},clip]{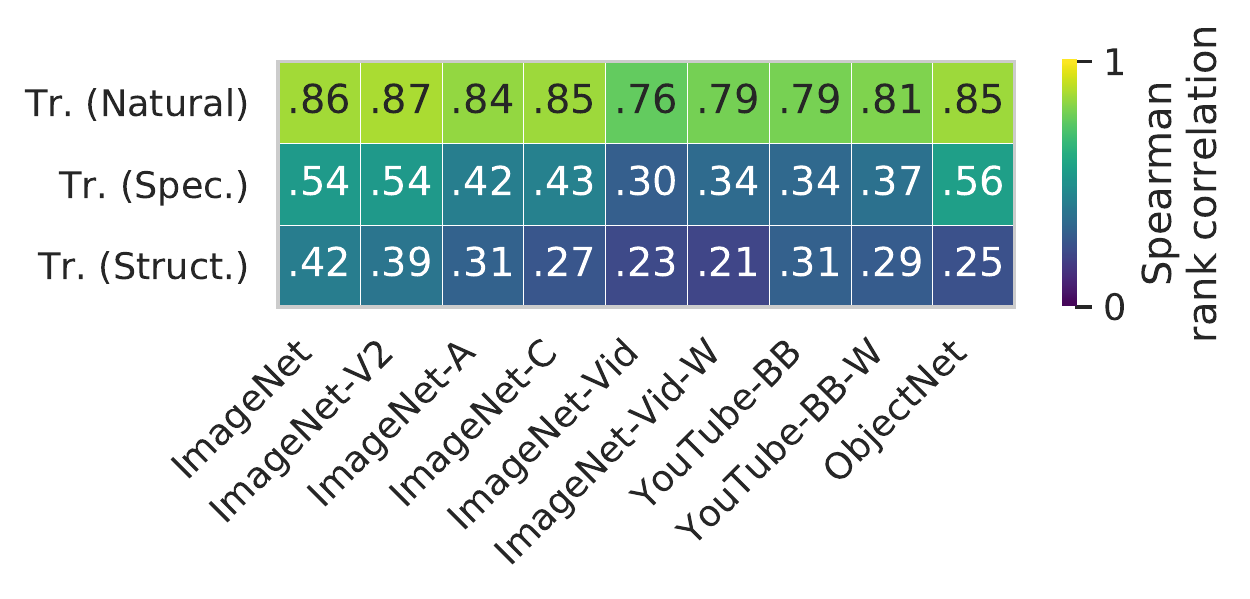}
    \caption{\footnotesize{
    The relationship between transfer learning, \textsc{ImageNet}, and robustness performance.
    \textbf{(Left)} Average score on all transfer benchmarks versus \textsc{ImageNet} performance.
    \textbf{(Center)} Average score on all robustness benchmarks versus average transfer performance.
    \textbf{(Right)} Correlation between different groups of transfer datasets (natural, specialized, structured), and robustness metrics.
    }\vspace{-5mm}}
    \label{fig:rt}
\end{tabular}
\end{figure*}
\section{A meta-analysis of robustness and transferability metrics}\label{sec:meta}

While many robustness metrics have been proposed to capture different sources of brittleness, it is not well understood how these metrics relate to each other.
We investigate the practical question of how useful the various metrics are in guiding design choices.
Further, we empirically analyze the relationship between robustness and transferability metrics, which is lacking in the literature, despite their close relationship. 
To analyze these questions, we evaluated 39 different models over 23 robustness metrics and the 19 transfer tasks.

\noindent\textbf{Metrics}\quad
\looseness-1For robustness, we measure the model accuracy on the \textsc{ImageNet}, \textsc{ImageNet-v2} (the \emph{matched frequency} variant) and \textsc{ObjectNet} datasets.
We also consider video datasets, \textsc{ImageNet-Vid} and \textsc{YouTube-BB}; we use both the accuracy metric and the \texttt{pm}-$10$ metric (suffix \textsc{-W}).
On \textsc{ImageNet-C} we report the AlexNet-accuracy-weighted \cite{krizhevsky2012imagenet} accuracy over all corruption times (called \emph{mean corruption error} in \cite{hendrycks2018benchmarking}).
To evaluate the transferability of the models, we use the VTAB-1K benchmark introduced in \Cref{sec:background}.
We evaluate average transfer performance across all 19 datasets, with 1000 examples each, as well as per-group performance.
To transfer a model we performed a sweep over two learning rates and schedules.
We report the median testing accuracy over three fine-tuning runs with parameters selected using a 800-200 example train-validation split.

\noindent\textbf{Models}\quad
To perform this meta-analysis we consider several model families.%
We evaluate ResNet-50 \cite{he2016deep} and six EfficientNet (B0 through B5) models \cite{tan2019efficientnet} including variants using AutoAugment \cite{cubuk2019autoaugment} and AdvProp \cite{xie2019adversarial}, which have been trained on \textsc{ImageNet}.
We include self-supervised SimCLR \cite{simclr} (variants: linear classifier on fixed representation (lin), fine-tuned on 10\% (ft-10), and 100\% (ft-100) of the \textsc{ImageNet} data), and self-supervised-semi-supervised (S4L)~\cite{s4l} models that have been fine-tuned to 10\% and 100\% of the \textsc{ImageNet} data.
We also consider a set of models that use other data sources.
Specifically, three NoisyStudent \cite{noisystudent} variants which use \textsc{ImageNet} and unlabelled data from the JFT dataset, BiT (BigTransfer)~\cite{bit} models that have been first trained on \textsc{ImageNet}, \textsc{ImageNet-21k}, or JFT and then transferred to \textsc{ImageNet} by fine-tuning, and the Video-Induced Visual Invariance (VIVI) model \cite{vivi}, which uses \textsc{ImageNet} and unlabelled videos from the YT8M dataset \cite{abu2016youtube}.
Finally, we consider the BigBiGAN \cite{bigbigan} model which has been first trained as a class-conditional generative model and then fine-tuned as an \textsc{ImageNet} classifier.
All details can be found in \Cref{app:model_overview}.

\begin{figure*}[t!]
    \centering
    \includegraphics[width=\linewidth]{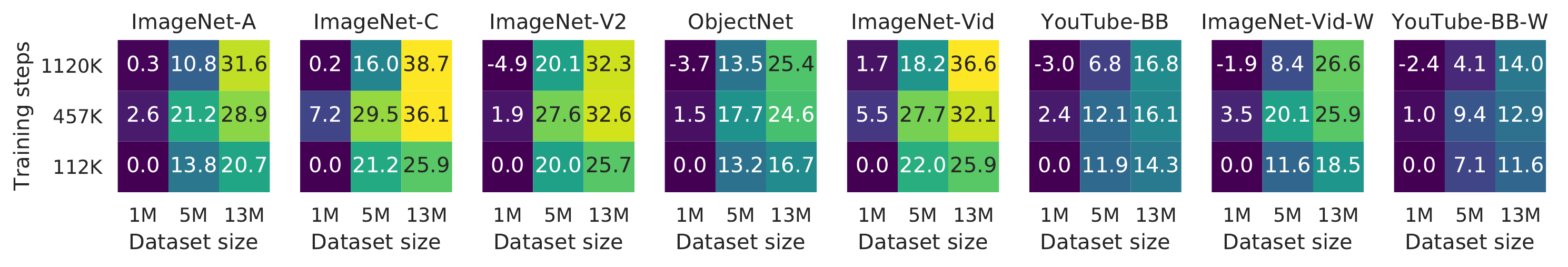}
    \includegraphics[width=\linewidth]{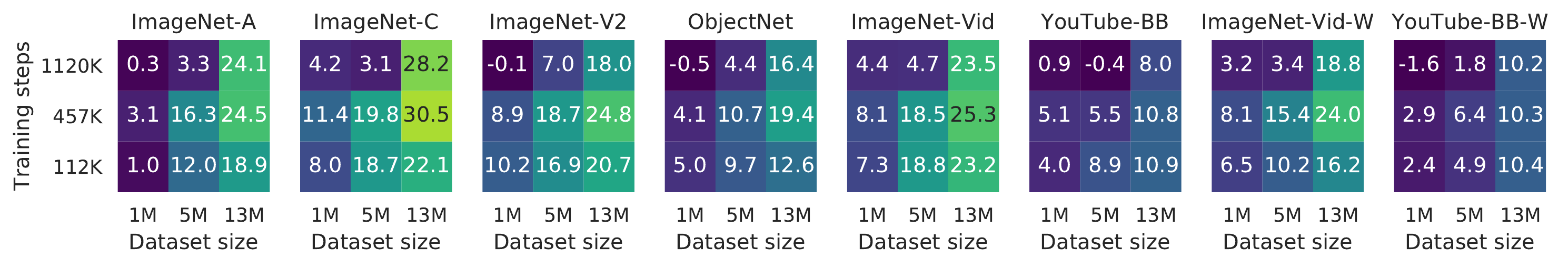}
    \vspace{-5mm}
    \caption{\textbf{(Top)}  Reduction (in \%) in classification error relative to the classification error of the model trained for 112k steps on 1M examples (bottom left corner) as a function of training iterations and training set size. The results are for a ResNet-101x3 trained on \textsc{ImageNet-21k} subsets. \textbf{(Bottom)} Relative reduction (in \%) in classification error going from a ResNet-50 to a ResNet-101x3 as a function of training steps and training set size (\textsc{ImageNet-21k} subsets). The reduction generally increases with the training set size and longer training. Hence, the right \emph{scaling laws} not only lead to in-distribution improvements, but also to simultaneous improvements across a heterogeneous set of OOD benchmarks. We investigate why these larger models achieve stronger performance across all benchmarks in Section~\ref{sec:synth}. 
    }
    \label{fig:model_size}
\end{figure*}

\begin{figure}[b]
    \centering
    \includegraphics[width=0.9\textwidth]{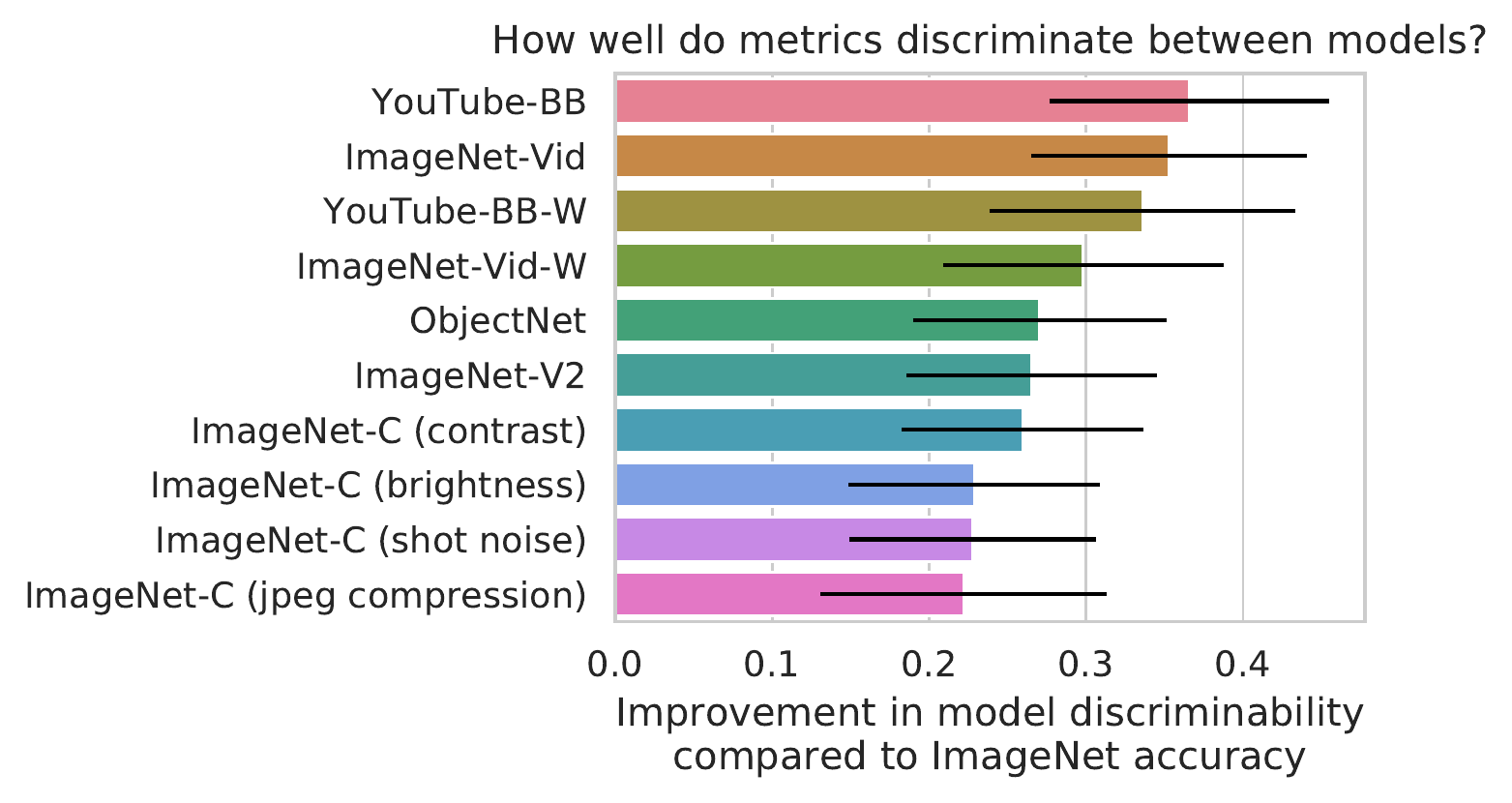}
    \caption{Informativeness of robustness metrics. Values indicate the difference in accuracy of a logistic classifier trained to discriminate between model types based on \textsc{ImageNet} accuracy plus one additional metric, compared to a classifier trained only on \textsc{ImageNet} accuracy (higher is better, top 10 metrics shown). Bars show mean$\pm$s.d.\ of 1000 bootstrap samples from the 39 models.}
    \label{fig:metric_discr}
\end{figure}

\vspace{1mm}
\noindent\textbf{How informative are robustness metrics for discriminating between models?}\quad
The goal of a metric is to discriminate between different models and thus guide design choices.
We therefore quantify the usefulness of each metric in terms of how much it improves the discriminability between the various models beyond the information provided by \textsc{ImageNet} accuracy.
Specifically, we train logistic regression classifiers to discriminate between the 12 model groups outlined above. We compared the performance of a classifier using only \textsc{ImageNet} accuracy as input feature, to a classifier using \textsc{ImageNet} and up to two of the other metrics, see Fig.\ \ref{fig:metric_discr} and \Cref{app:analysis}. 
We found that most of the tested metrics provide little increase in model discriminability over \textsc{ImageNet} accuracy.
We further, similarly to \cite{taori2020measuring}, found that all metrics are highly rank-correlated with each other, which we present in \Cref{app:analysis}.
Of course, these results are conditioned on the size and composition of our dataset, and may differ for a different set of models.
However, based on our collection of 39 models in 12 groups, the most informative metrics are those based on different datasets and/or video, rather than \textsc{ImageNet}-derived datasets. 

\vspace{1mm}
\noindent\textbf{How related are OOD robustness and transfer metrics?}\quad
\looseness-1Next, we turn to transfer learning.
It has been observed that better \textsc{ImageNet} models transfer better \cite{DBLP:conf/cvpr/KornblithSL19, zhai2019largescale}.
Since robustness metrics correlated strongly with \textsc{ImageNet} accuracy, we might expect a similar relationship.
To get an overall view, we compute the mean of all robustness metrics, and compare it to transfer performance.
\Cref{fig:rt} (center) shows this average robustness plotted against transfer performance, while \Cref{fig:rt} (left) shows transfer versus \textsc{ImageNet} accuracy.
Indeed, we observe a large correlation coefficient $\rho=0.73$ between robustness and transfer metrics; however, the correlation is not stronger than between transfer and \textsc{ImageNet}.
Further, we compute the correlation of the residual robustness score (mean robustness minus \textsc{ImageNet} accuracy) against transfer score, and find only a weak relationship of $\rho=0.12$.
This indicates that robustness metrics, on aggregate, do not provide additional signal that predicts model transferability beyond that of the base \textsc{ImageNet} performance.
We do, however, see some interesting differences in the relative performances of different model groups.
Certain model groups, while attaining reasonable \textsc{ImageNet}/robustness scores, transfer less well to VTAB.
Therefore, \emph{there are factors unrelated to robust inference that do influence transferability}.
One example is batch normalization which is outperformed by group normalization with weight standardization in transfer~\cite{bit}.
Next, we break down the correlation by robustness metrics and transfer datasets in Fig.\ \ref{fig:rt} (right).
We see that each metric correlates similarly with the task groups.
However, for the groups that require more distant transfer (Specialized, Structured), no metric predicts transferability well.
Perhaps surprisingly, raw \textsc{ImageNet} accuracy is the best predictor of transfer to \emph{structured} tasks, indicating that robustness metrics do not relate to challenging transfer tasks, at least not more than raw \textsc{ImageNet} accuracy.

\vspace{1mm}
\noindent\textbf{Summary}\quad
Metrics based on ImageNet have very little additional discriminative power over ImageNet accuracy, while those not based on ImageNet have more, but their additional discriminative power is still low---popular robustness metrics provide marginal complementary information. Transferability is also related to \textsc{ImageNet} accuracy, and hence robustness.
We observe that while there is correlation, transfer highlights failures that are somewhat independent of robustness.
Further, no particular robustness metric appears to correlate better with any particular group of transfer tasks than \textsc{ImageNet} does. Inspired by these results, we next investigate strategies known to be effective for \textsc{ImageNet} and transfer learning on the OOD robustness benchmarks.

\section{Scaling laws for OOD performance}\label{sec:designchoices}
Increasing the scale of pre-training data, model architecture, and training steps have recently led to diminishing improvements in terms of \textsc{ImageNet} accuracy. By contrast, it has been recently established that scaling along these axes can lead to substantial improvements in transfer learning performance \cite{bit, tan2019efficientnet}. In the context of robustness, this type of scaling has been explored less. While there are some results hinting that scale can improve robustness \cite{hendrycks2018benchmarking, recht2019imagenet,  noisystudent, touvron2020fixing}, no principled study decoupling the different scale axes has been performed. Given the strong correlation between transfer performance and robustness, this motivates the systematic investigation of the effects of the pre-training data size, model architecture size, training steps, and input resolution. While paramount to the out-of-distribution performance, as we find, these pretraining design choices have not yet received a great deal of attention from the community.

\vspace{1mm}
\noindent\textbf{Setup}\quad We consider the standard \textsc{ImageNet} training setup~\cite{he2016deep} as a baseline, and scale up the training accordingly.
To study the impact of dataset size, we consider the \textsc{ImageNet-21k} \cite{deng2009imagenet} and \textsc{JFT} \cite{DBLP:conf/iccv/SunSSG17} datasets for the experiments, as pre-training on either of them has shown great performance in transfer learning~\cite{bit}. We scale from the \textsc{ImageNet} training set size ($1.28$M images) to the \textsc{ImageNet-21k} training set size (13M images, about $10$ times larger than \textsc{ImageNet}).
To explore the effect of the model size, we use a ResNet-50 as well as the deeper and $3 \times$wider ResNet-101x3 model.
We further investigate the impact of the training schedule as larger datasets are known to benefit from longer training for transfer learning \cite{bit}.
To disentangle the impact of dataset size and training schedules, we train the models for every pair of dataset size and schedule. 

We fine-tune the trained models to \textsc{ImageNet} using the BiT HyperRule~\cite{bit}, and assess their OOD generalization performance in the next section.
Throughout, we report the reduction in classification error relative to the model which was trained on the smallest number of examples and for the fewest iterations, and which hence achieves the lowest accuracy. Other details are presented in \Cref{app:large-scale}.

\vspace{1mm}
\noindent\textbf{Pre-training dataset size impact}\quad The results for the ResNet-101x3 model are presented in Fig.\ \ref{fig:model_size}. When pre-trained on \textsc{ImageNet-21k}, the OOD classification error significantly decreases with increasing pre-training dataset size and duration: We observe relative error reductions of $20$-$30\%$ when going from  112k steps on 1M data points to 1.12M steps on 13M data points. The reductions are least pronounced for \textsc{YouTube-BB(-W)}. Note that training for $1.12$M steps leads to a lower accuracy than training for only 457k steps unless the full \textsc{ImageNet-21k} dataset is used. For models trained on \textsc{JFT} we observe a similar behavior except that training for 1.12M steps often leads to a higher accuracy than training for 457k steps even when only 1M or 5M data points are used (c.f.\ Appendix~\ref{app:large-scale}).
These results suggest that, if the models have enough capacity, increasing the amount of pre-training data, without any additional changes, leads to substantial gains in all datasets \emph{simultaneously} which is in line with recent results in transfer learning~\cite{bit}.

\vspace{1mm}
\noindent\textbf{Model size impact}\quad \Cref{fig:model_size} shows the relative reduction in classification error when using a ResNet-101x3 instead of a ResNet-50 as a function of the number of training steps and the dataset size. It can be seen that increasing the model size can lead to substantial reductions of $5$--$20\%$. For a fixed training duration, using more data always helps. However, on \textsc{ImageNet-21k}, training too long can lead to increases in the classification error when the model size is increased, unless the full \textsc{ImageNet-21k} is used. This is likely due to overfitting. This effect is much less pronounced when \textsc{JFT} is used for training. \textsc{JFT} results are presented in Appendix~\ref{app:large-scale}. Again, reductions in classification error are least pronounced for \textsc{YouTube-BB/YouTube-BB-W}. 

\begin{figure}[t]
    \centering
    \includegraphics[width=\textwidth]{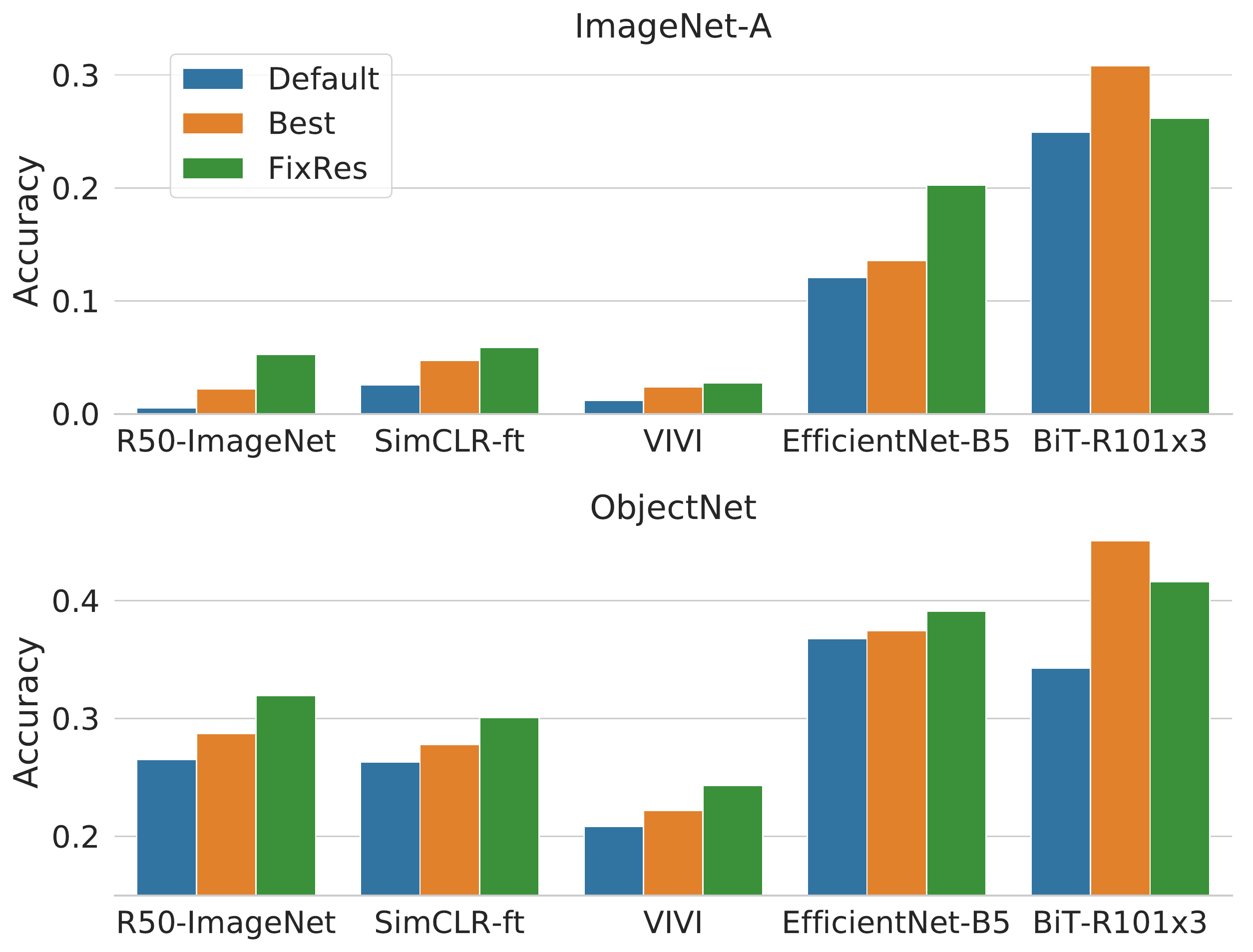}
    \caption{Comparison of different types of evaluation preprocessing and resolutions. (Default, blue): Accuracy obtained for the preprocessing and resolution proposed by the authors of the respective models. (Best, orange): The accuracy when selecting the best resolution from $\{64, 128, 224, 288, 320, 384, 512, 768\}$. (FixRes, green): Applying FixRes for the same set of resolutions and selecting the best resolution. Increasing the evaluation resolution and additionally using FixRes helps across a large range of models and pretraining datasets on \textsc{ImageNet-A} and \textsc{ObjectNet}.\vspace{-6mm}}
    \label{fig:fixres}
\end{figure}

\vspace{1mm}
\noindent\textbf{Testing resolution and OOD robustness}\quad
During training, images are typically cropped randomly, with many crop sizes and aspect ratios, to prevent overfitting. In contrast, during testing, the images are usually rescaled such that the shorter side has a pre-specified length, and a fixed-size center crop is taken and then fed to the classifier. This leads to a mismatch in object sizes between training and testing. Increasing the resolution at which images are tested leads to an improvement in accuracy across different architectures~\cite{touvron2019fixing, touvron2020fixing}. Furthermore, additional benefits can be obtained by applying \emph{FixRes} --- fine-tuning the network on the \emph{training} set with the test-time preprocessing (i.e.\ omitting random cropping with aspect ratio changes), and at a higher resolution. We explore the effect of this discrepancy on the robustness of different architectures. As some of the robustness datasets were collected differently from \textsc{ImageNet}, discrepancies in the cropping are likely. We investigate both adjusting test-time resolution and applying FixRes. For FixRes, we use a simple setup with a single schedule and learning rate for all models (except using a $10\times$ smaller learning rate for the BiT models), and without heavy color augmentation as in \cite{touvron2019fixing} or label smoothing as in \cite{touvron2020fixing}. We did not extensively tune hyperparameters, but chose a setup that works reasonably well across architectures and training datasets.
Note that changing the resolution can be seen as scaling the computational resources available to the model, as both training and inference costs grow with the resolution.

Following the protocol of the FixRes paper \cite{touvron2019fixing}, we evaluate each model for all resolutions in $\{64, 128, 224, 288, 320, 384, 512, 768\}$ to \emph{illustrate the potential of adapting the testing resolution} (in practice we do not have access to an OOD validation set so we cannot select the optimal solution in advance). For conciseness, we show the accuracy for \textsc{ImageNet-A} and \textsc{ObjectNet} at the testing resolution proposed by the authors of the respective architecture along with the highest accuracy across testing resolutions (\Cref{fig:fixres}). The results for other datasets and resolutions are deferred to \Cref{app:resolution}.

We start by discussing observations that apply to most models, excluding the BiT models which will be discussed below. While FixRes only leads to marginal benefits on \textsc{ImageNet}, it can lead to substantial improvements on the robustness metrics.
Choosing the optimal testing resolution leads to a significant increase in accuracy on \textsc{ImageNet-A} and \textsc{ObjectNet} in most cases, and applying FixRes often leads to additional substantial gains. For \textsc{ObjectNet}, fine-tuning with testing preprocessing (i.e.\ fine-tuning with central cropping instead of random cropping as used during training) can help even without increasing resolution.

\begin{figure*}[t]
    \centering
    \begin{subfigure}[b]{0.49\textwidth}
        \centering
        \includegraphics[width=0.3\linewidth]{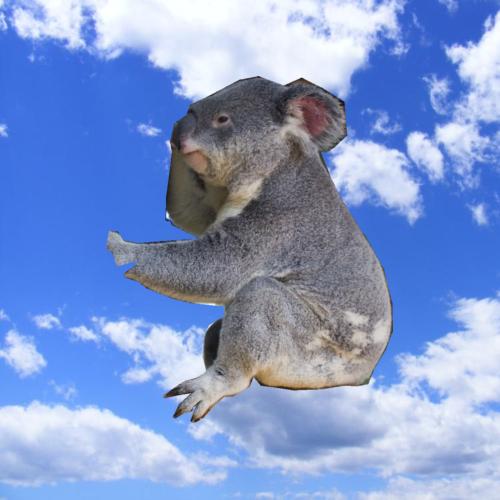}
        \includegraphics[width=0.3\linewidth]{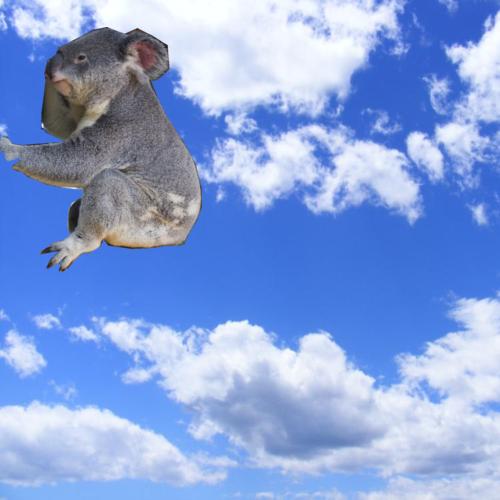}
        \includegraphics[width=0.3\linewidth]{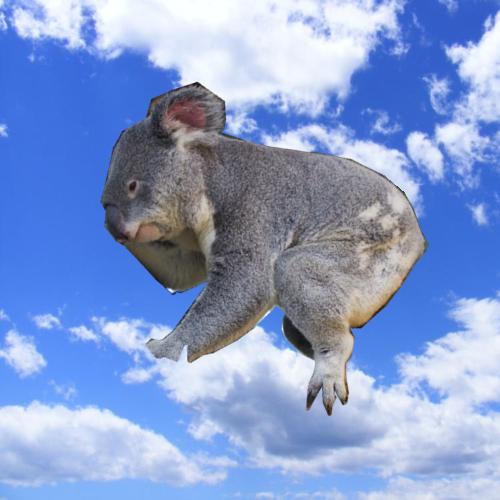}\\
        \includegraphics[width=0.3\linewidth]{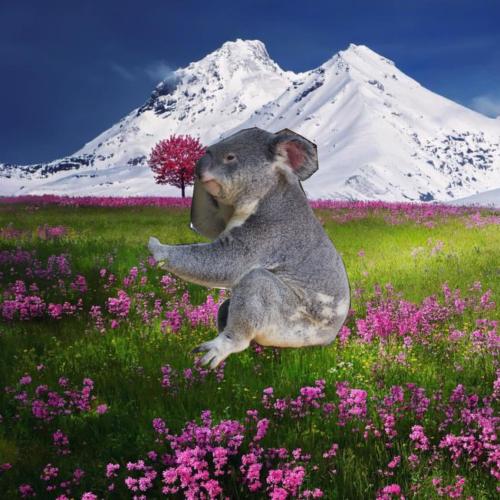}
        \includegraphics[width=0.3\linewidth]{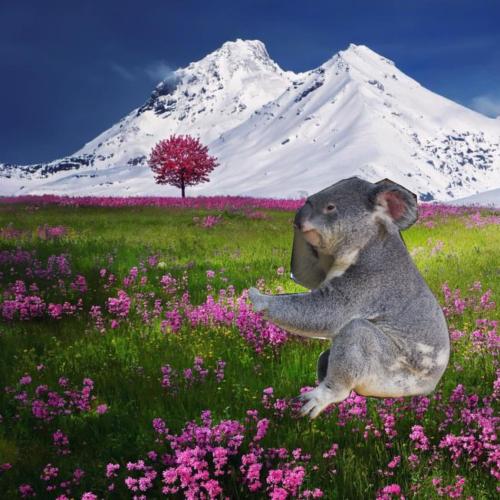}
        \includegraphics[width=0.3\linewidth]{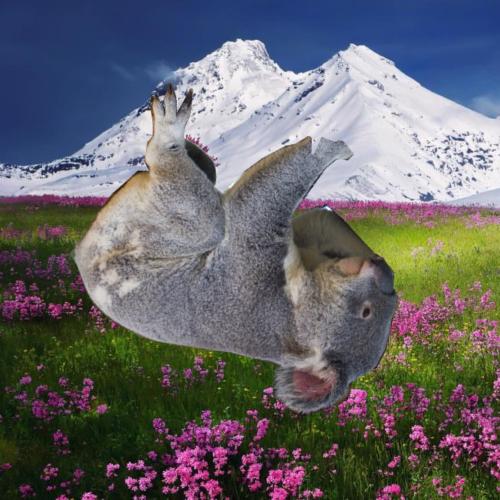}  
        \end{subfigure}\hfill
        \begin{subfigure}[b]{0.49\textwidth}  
            \centering 
  \includegraphics[width=.85\linewidth]{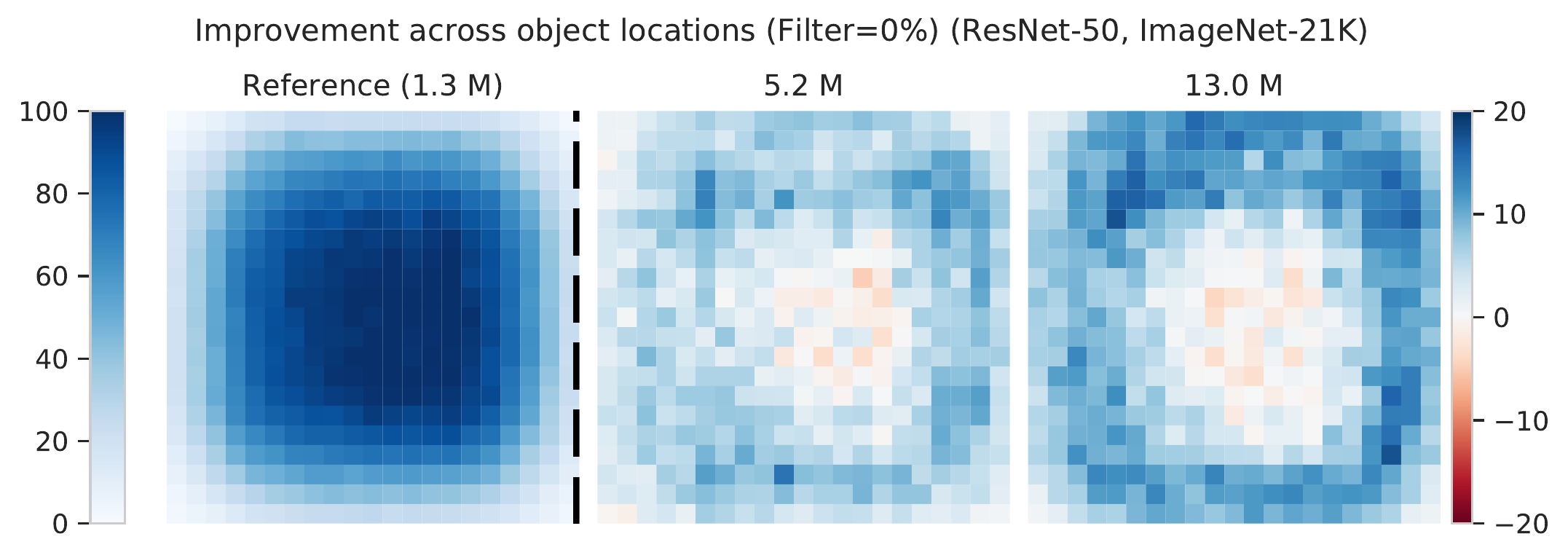}
  \includegraphics[width=.85\linewidth]{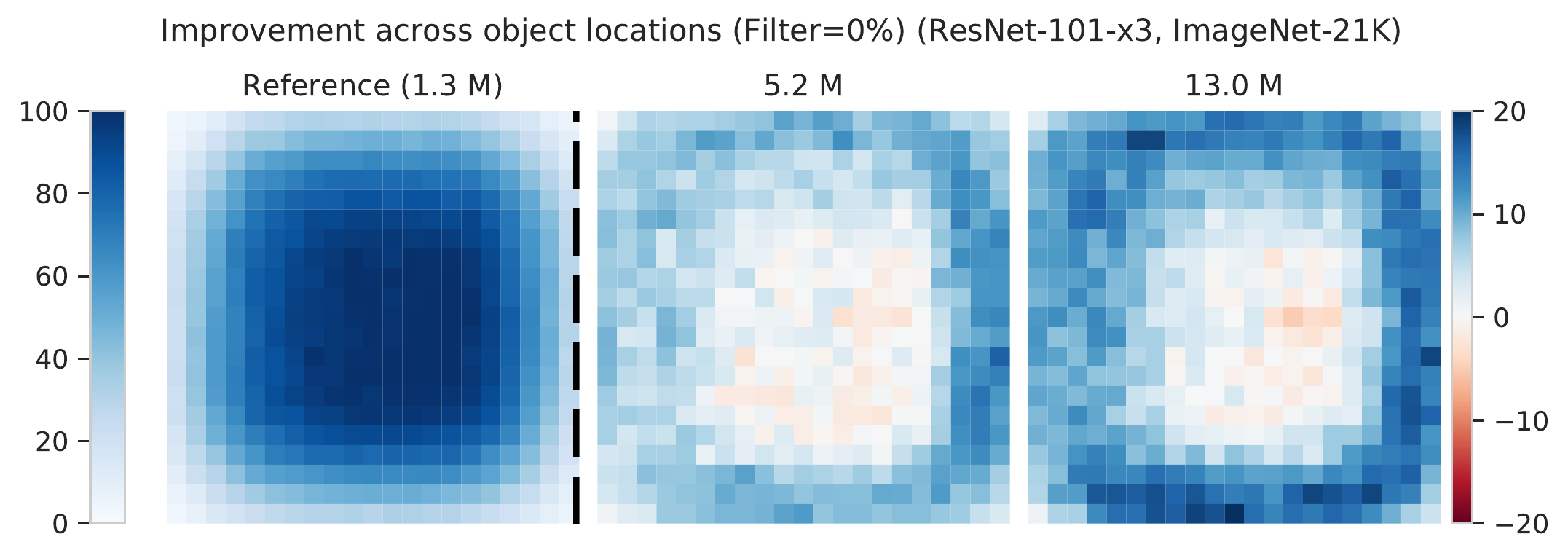}  
        \end{subfigure}
        \caption{\textbf{(Left)} Sample images from our synthetic dataset. We consider 614 foreground objects from 62 classes and 867 backgrounds and vary the object location, rotation angle, and object size for a total of 
        \num{611608} images. \textbf{(Right)} In the first column, for each location on the grid, we compute the average accuracy. Then, we normalize each location by the 95\textsuperscript{th} percentile across all locations, which quantifies the gap between the locations where the model performs well, and the ones where it under-performs (first column, dark blue versus white). Then, we consider models trained with more data, compute the same normalized score, and plot the \emph{difference} with respect to the first column. We observe that, as dataset size increases, sensitivity to object location decreases -- the outer regions improve in relative accuracy more than the inner ones (e.g. dark blue vs white on the second and third columns). The effect is more pronounced for the larger model. The full set of results is presented in Figure~\ref{fig:synth-location-0pc} in Appendix~\ref{app:synth}.\label{fig:synth_scale_location} \vspace{-2mm}}
    \end{figure*}

Increasing the resolution and/or applying FixRes often slightly helps on \textsc{ImageNet-V2}. 
For \textsc{ImageNet-C}, the optimal testing resolution often corresponds to the resolution used for training, and applying FixRes rarely changes this picture. This is not surprising as the \textsc{ImageNet-C} images are cropped to 224 pixels by default, and increasing the resolution does not add any new information to the image.
For the video-derived robustness datasets \textsc{ImageNet-Vid-Robust} and \textsc{YouTube-BB-Robust}, evaluating at a larger testing resolution and/or applying FixRes at a higher resolution can substantially improve the accuracy on the anchor frame and the robustness accuracy for small EfficientNet and ResNet models, but does not help the larger ones.
For the BiT models, the resolution suggested by the authors is almost always optimal, except on \textsc{ObjectNet} and \textsc{ImageNet-A}, where changing the preprocessing considerably helps. FixRes arguably does not lead to improvements as it was already applied in BiT as a part of the BiT HyperRule. 

\vspace{1mm}
\noindent\textbf{Summary}\quad
These empirical results point to the following conclusion: for models with enough capacity, increasing the amount of pre-training data, with no additional changes, leads to substantial gains in all considered OOD generalization tasks \emph{simultaneously}. Secondly, resolution adjustments as outlined above can address the considerable distribution shift caused by resolution mismatch.

\section{\textsc{SI-Score}: A fine-grained analysis of robustness to common factors of variation}\label{sec:synth}
The results in Section~\ref{sec:designchoices} do not reveal the underlying reasons for the success of larger models trained on more data on all robustness metrics. Intuitively, one would expect that these models are more invariant to specific factors of variation, such as object location, size, and rotation. However, a systematic assessment hinges on testing data which can be varied according to these axes in a controlled way. At the same time, the combinatorial nature of the problem precludes any large-scale systematic data collection scheme. 

In this work we present a scalable alternative and construct a novel synthetic dataset for fine-grained evaluation: \textsc{SI-Score} (Synthetic Interventions on Scenes for Robustness Evaluation). In a nutshell, we paste a large collection of objects onto uncluttered backgrounds (\Cref{fig:synth_scale_location}, \Cref{fig:synth-examples-appendix}), and can thus conduct controlled studies by systematically varying the object class, size, location, and orientation.\footnote{The synthetic dataset and code used to generate the dataset are open-sourced on \href{https://github.com/google-research/si-score}{GitHub and are being hosted by the Common Visual Data Foundation}.}
\begin{table}[t]
    \centering
    \footnotesize{
     \begin{tabular}{p{0.125\linewidth} p{0.625\linewidth} p{0.1\linewidth}} \\ 
     \toprule
\textsc{\textsc{F.O.V.}}	 & \textsc{Dataset Configuration} & \textsc{Images} \\ \midrule
\textsc{Size} & Objects upright in the center, sizes from 1\% to 100\% of the image area in 1\% increments.& \num{92884}\\
\textsc{Location} & Objects upright. Sizes are 20\% of the image area. We do a grid search of locations, dividing the x-coordinate and y-coordinate dimensions into 20 equal parts each, for a total of 441 coordinate locations. & \num{479184}\\
\textsc{Rotation} & Objects in the center, sizes equal to 20\%, 50\%, 80\% or 100\% of the image size. Rotation angles ranging from 1 to 341 degrees counterclockwise in 20-degree increments.& \num{39540}\\
     \toprule
     \end{tabular}
     }
    \caption{Synthetic dataset details. The first column shows the relevant factor of variation (F.O.V.). When there are multiple values for multiple factors of variation, we generate the full cross product of images.} \label{table:synth-data-description}
\end{table}

\vspace{2mm}
\noindent\textbf{Synthetic dataset details}\quad The foregrounds were extracted from OpenImages~\cite{OpenImages} using the provided segmentation masks. We include only object classes that map to ImageNet classes. We also removed all objects that are tagged as occluded or truncated, and manually removed highly incomplete or inaccurately labeled objects. The backgrounds were images from nature taken from \emph{pexels.com} (the license therein allows one to reuse photos with modifications). We manually filtered the backgrounds to remove ones with prominent objects, such as images focused on a single animal or person. In total, we converged to 614 object instances across 62 classes, and a set of 867 backgrounds.

We constructed three subsets for evaluation, one corresponding to each factor of variation we wanted to investigate, as shown in \Cref{table:synth-data-description}. In particular, for each object instance, we sample two backgrounds, and for each of these object-background combinations, we take a cross product over all the factors of variation. For the datasets with multiple values for more than one factor of variation, we take a cross product of all the values for each factor of variation in the set (object size, rotation, location). For example, for the rotation angle dataset, there are four object sizes and 18 rotation angles, so we do a cross product and have 72 factor of variation combinations. For the object size and rotation datasets, we only consider images where objects are at least 95\% in the image. For the location dataset, such filtering removes almost all images where objects are near the edges of the image, so we do not do such filtering. Note that since we use the central coordinates of objects as their location, at least 25\% of each object is in the image even if we do not do any filtering. The results in the following sections are similar when filtering out objects that are less than 50\% or 75\% in the image. 

\begin{figure*}[t]
    \centering
    \begin{subfigure}[b]{0.49\textwidth}   
        \centering 
\includegraphics[width=.99\linewidth]{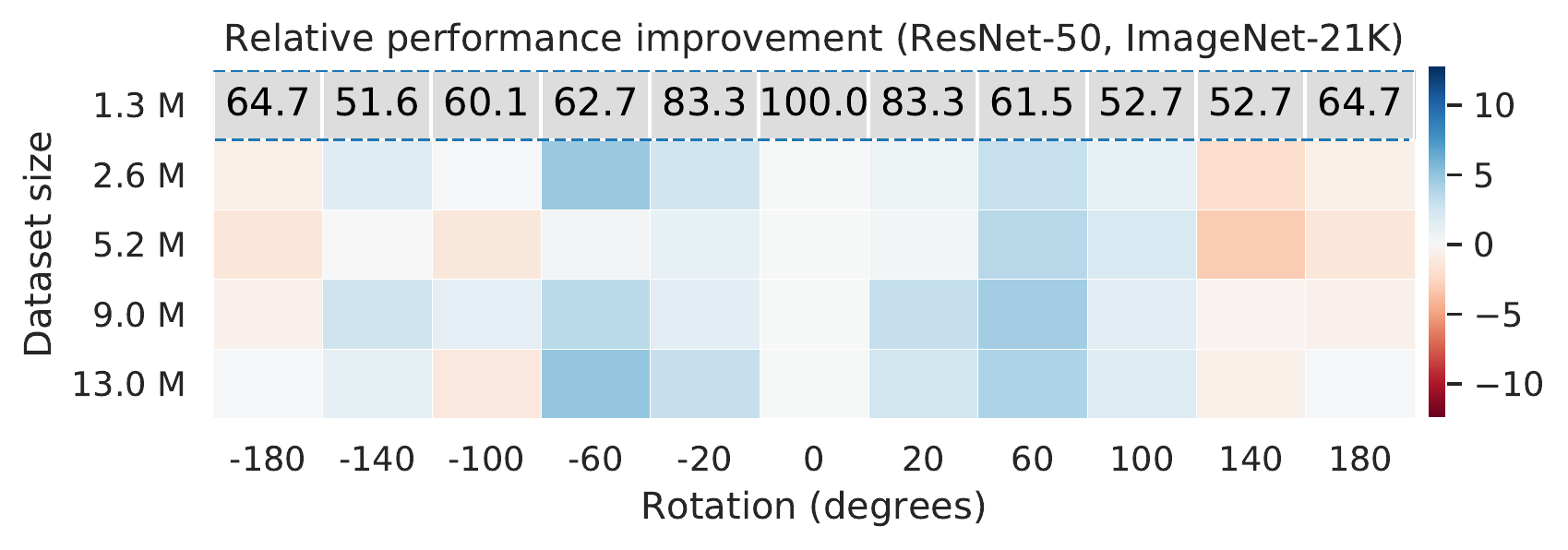}
\includegraphics[width=.99\linewidth]{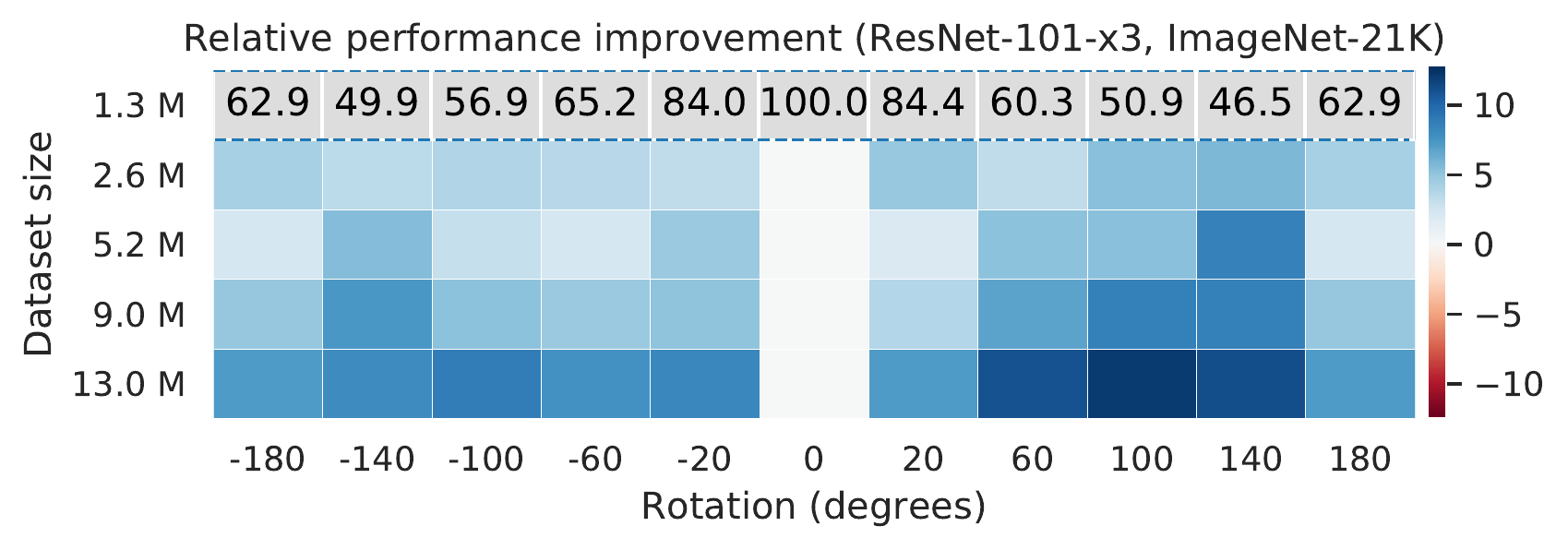}
    \end{subfigure}\hfill
    \begin{subfigure}[b]{0.49\textwidth}
        \setlength{\abovecaptionskip}{5pt} %
        \centering 
          \includegraphics[width=1.0\linewidth]{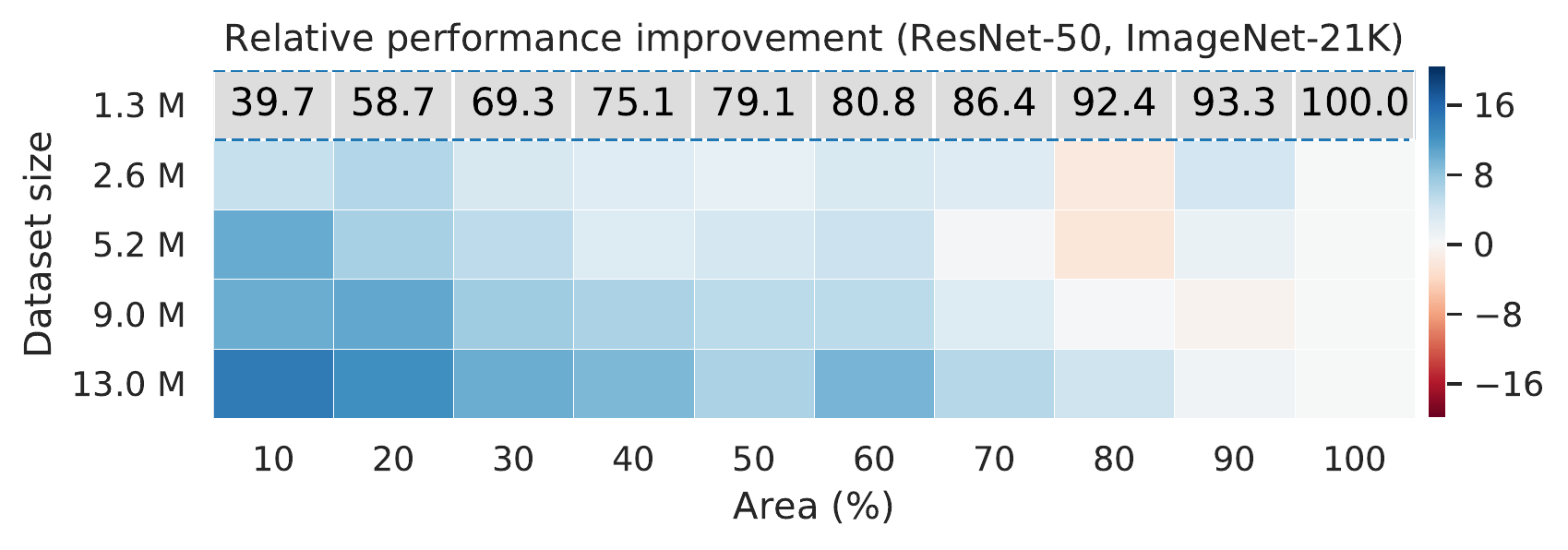}
          \includegraphics[width=1.0\linewidth]{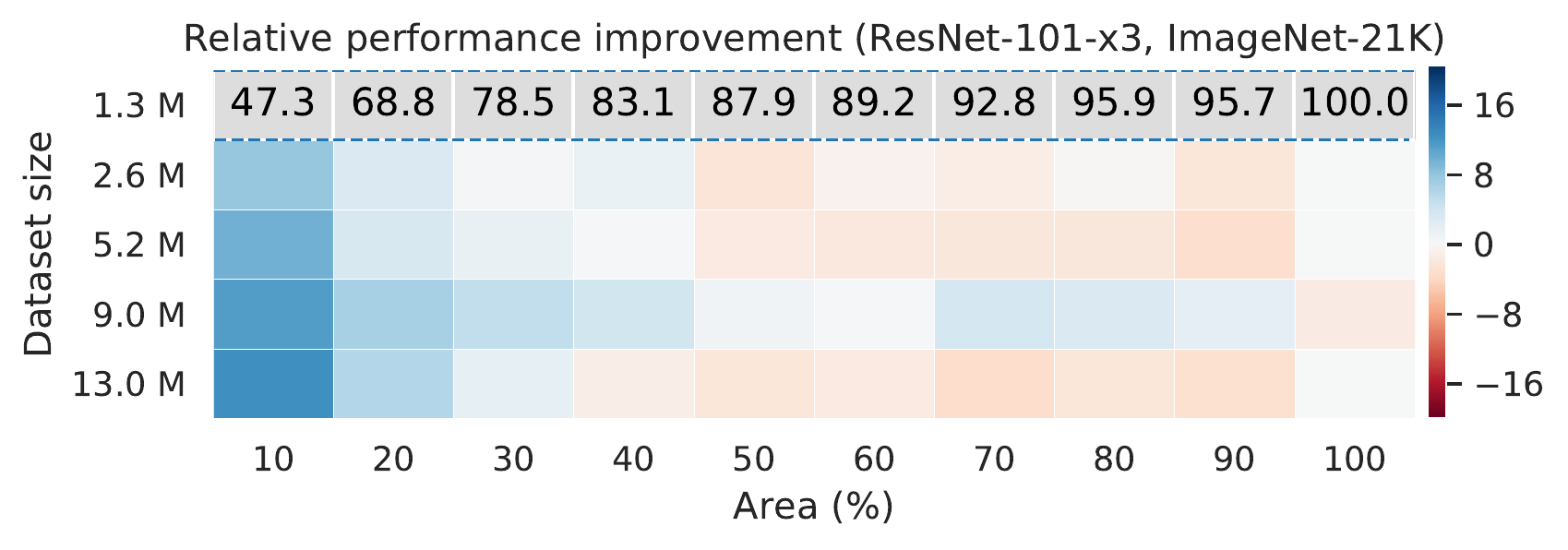}
          \vspace{-4mm}
    \end{subfigure}
    \vspace{-2mm}
    \caption[]
    {
    \textbf{(Left)}
    In the first row of both plots we show the ratio of the accuracy and the best accuracy (across all rotations). For the second row (model trained on 2.6M instances) and other rows, we compute the same normalized score and visualize the difference with the first row. Larger positive differences with the first row imply a more uniform behavior across object rotations. We observe that, as the dataset size increases, the average prediction accuracy across various rotation angles becomes more uniform. The effect is more pronounced for the larger model.
    \textbf{(Right)}
    Similarly, the average accuracy across various object sizes becomes more uniform for both models. As expected, the improvement is most pronounced for small object sizes covering $10$--$20\%$ of the pixels. The full set of results is presented in Figures~\ref{fig:synth-area} and \ref{fig:synth-rotation} in Appendix~\ref{app:synth}.}
    \label{fig:synth_scale_rotation}
\end{figure*}

\noindent\textbf{Learned invariances as a function of scale}\quad
We study one factor of variation at a time. For example, when studying the impact of changing the location of the object center, we measure the average performance for each location over a uniform grid. Building on our investigation in the previous section, we test whether increasing model size and dataset size improves robustness to these three factors of variation by evaluating the ResNet-50 and ResNet-101x3 models. We observe that the models indeed become more invariant to object location (\Cref{fig:synth_scale_location}), rotation (\Cref{fig:synth_scale_rotation}, left), and size (\Cref{fig:synth_scale_rotation}, right) as the pre-training set size increases. Specifically, as we pre-train on more data, the average prediction accuracy across various object locations, sizes, and rotation angles becomes more uniform. Furthermore, the larger ResNet-101x3 model is indeed more robust. Analogous results on the \textsc{JFT} dataset are presented in Appendix~\ref{app:synth}.

\section{Related work}\label{sec:relate-work}
There has been a growing literature exploring the robustness of image classification networks. 
Early investigations in face and natural image recognition found that performance degrades by introducing blur, Gaussian noise, occlusion, and compression artifacts, but less by color distortions~\cite{dodge2016understanding,karahan2016image}.
Subsequent studies have investigated brittleness to similar corruptions~\cite{roy2018effects,zhou2017classification}, as well as to impulse noise~\cite{hosseini2017google}, photometric perturbations~\cite{temel2018cure}, and small shifts and other transformations~\cite{azulay2019deep,engstrom2020identifying,zhang2019making}.
CNNs have also been shown to over-rely upon texture rather than shape to make predictions, in contrast to human behavior~\cite{geirhos2018imagenet}.
Robustness to adversarial attacks \cite{goodfellow2014explaining} is a related, but distinct problem, where performance under worst-case perturbations are studied.
In this paper we did not study such adversarial robustness, but have focused on average-case robustness to natural perturbations.

2Several techniques have been shown to improve model robustness on these datasets.
Using better data augmentation can improve performance on data with synthetic noise~\cite{hendrycks2019augmix,lopes2019improving}.
Auxiliary self-supervision \cite{ssgan,s4l} can improve robustness to label noise and common corruptions \cite{hendrycks2019self}.
Transductive fine-tuning using self-supervision on the test data improves performance under distribution shift~\cite{sun2019test}.
Training with adversarial perturbations improves many robustness benchmarks if one uses separate Batch-Norm parameters for clean and adversarial data~\cite{xie2019adversarial}.
Finally, additional pre-training using very large auxiliary datasets has recently shown significant improvements in robustness.
Noisy Student~\cite{noisystudent} reports good performance on several robustness datasets, while Big Transfer (BiT)~\cite{bit} reports strong performance on the \textsc{ObjectNet} dataset \cite{barbu2019objectnet}.

Deep networks are often trained by pre-training the network on a different problem and then fine-tuning on the target task.  
This pre-training is often referred to as representation learning; 
representations can be trained using supervised~\cite{huh2016makes,bit}, weakly-supervised~\cite{mahajan2018exploring}, or unsupervised data~\cite{doersch2015unsupervised,bigbigan,vivi,noisystudent}.
Recent benchmarks have been proposed to evaluate transfer to several datasets, to assess generalization to tasks with different characteristics, or those disjoint from the pre-training data~\cite{triantafillou2019meta,zhai2019largescale}.
While state-of-the-art performance on many competitive datasets is attained via transfer learning~\cite{noisystudent,bit}, the implications for final robustness metrics remain unclear.

Creating synthetic datasets by inserting objects onto backgrounds has been used for training \cite{zhao2020distilling, dwibedicutpaste2017, ghiasi2020copypaste} and evaluating models~\cite{bit}, but previous works do not systematically vary object size, location or orientation, or analyze translation and rotation robustness only at the image level~\cite{engstrom2017}.

Given the lack of a consensus on what ``natural'' perturbations \emph{are},  there are no established general laws on how models behave under various data shifts.
Concurrently, \cite{taori2020measuring} investigated whether higher accuracy on synthetic datasets translates to superior performance on natural OOD datasets. They also identify model size and training data set size as the only technique providing a benefit.
In \cite{hendrycks2020many} the authors list several of the hypotheses that appear in the literature, and collect new datasets that provide (both positive and negative) evidence for their soundness.

\section{Limitations and future work}
We analyzed OOD generalization and transferability of image classifiers, and demonstrated that model and data scale together with a simple training recipe lead to large improvements.
However, these models do exhibit substantial performance gaps when tested on OOD data, and further research is required.
Secondly, this approach hinges on the availability of curated datasets and significant computing capabilities which is not always practical. Hence, we believe that transfer learning, i.e.\ train once, apply many times, is the most promising paradigm for OOD robustness in the short term. One limitation of this study is that we consider image classification models fine-tuned to the \textsc{ImageNet} label space which were developed with the goal of optimizing the accuracy on the \textsc{ImageNet} test set. While existing work shows that we do not overfit to \textsc{ImageNet}, it is possible that these models have correlated failure modes on datasets which share the biases with~\textsc{ImageNet}~\cite{recht2019imagenet}. This highlights the need for datasets which enable fine-grained analysis for all important factors of variation and we hope that our dataset will be useful for researchers.

The introduced synthetic data can be used to investigate other qualitative differences between models. For example, when comparing ResNet-50s trained on ImageNet, a ResNet using GroupNorm does better on smaller objects than one with BatchNorm, whereas the model with BatchNorm does better on larger objects (Figure~\ref{fig:synth-gn-bn} in the appendix). While a thorough investigation is beyond the scope of this work, we hope that \textsc{SI-Score} will be useful for such future studies.

Instead of requiring the model to work under various dataset shifts, one can ask an alternative question: assuming that the model will be deployed in an environment significantly different from the training one, can we at least quantify the model uncertainty for each prediction? This important property remains elusive for moderate-scale neural networks~\cite{DBLP:conf/nips/SnoekOFLNSDRN19}, but could potentially be improved by large-scale pretraining which we leave for future work.

{\small
\bibliographystyle{ieee_fullname}
\bibliography{main}
}

\clearpage

\appendix

\onecolumn

\section{Analysis of existing robustness and transfer metrics}
\label{app:analysis}

Here, we provide additional details related to the analyses and benchmarks presented in Section~\ref{sec:meta}.

\subsection{Robustness metric correlation}
\begin{figure}[ht]
    \centering
    \includegraphics[width=0.5\textwidth]{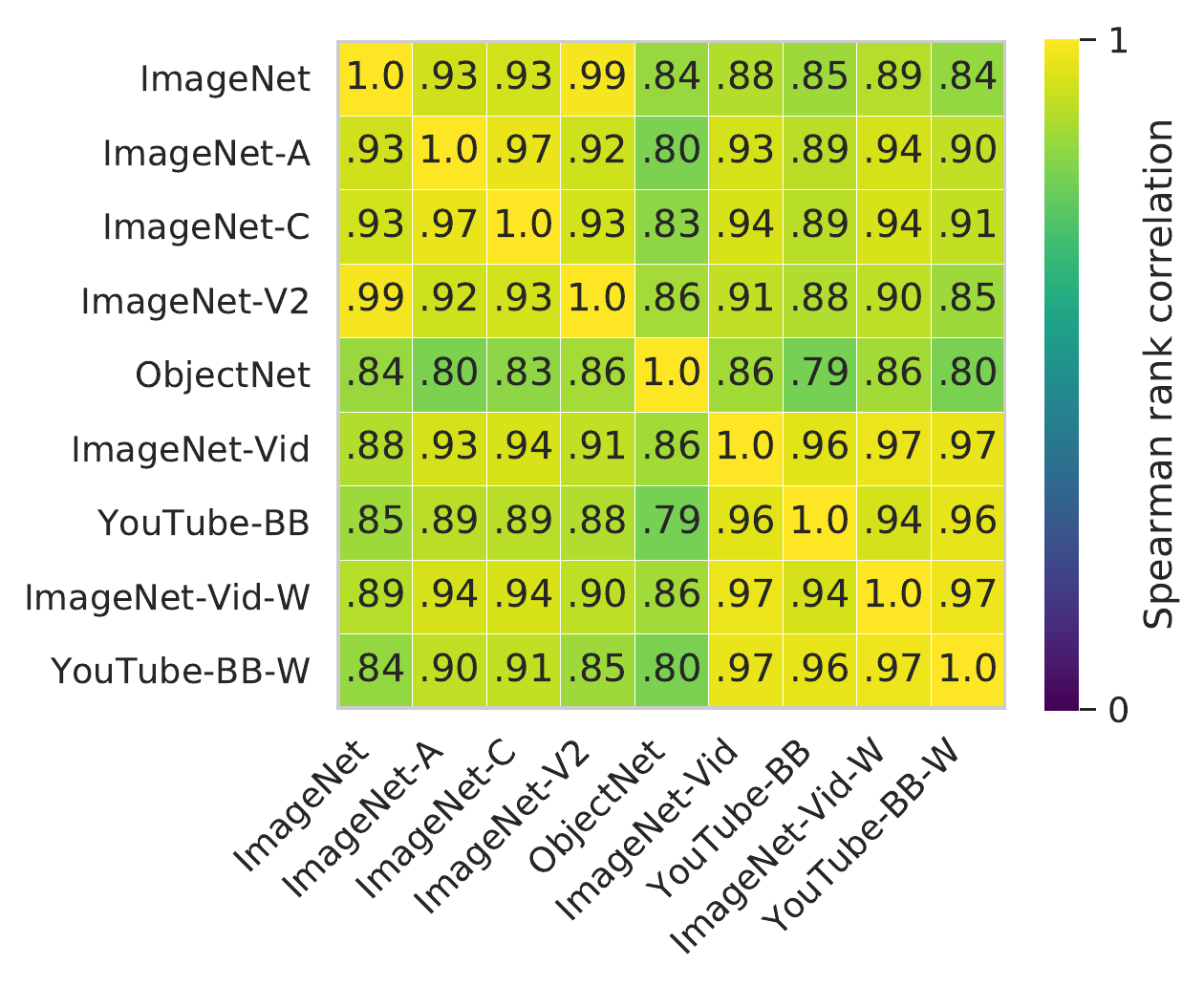}
    \caption{Spearman's rank correlation between accuracies on the eight robustness datasets. Samples were taken from 39 models across various model families presented in Table~\ref{tab:model_training_overview}.
    \label{fig:metric_correlations}}
\end{figure}

\subsection{Dimensionality of the space of robustness metrics}
\label{app:dimensionality_details}
\begin{figure*}[h]
    \begin{subfigure}[normal]{0.5\textwidth} 
    \centering
    \raisebox{0.0cm}{\includegraphics[width=0.7\textwidth]{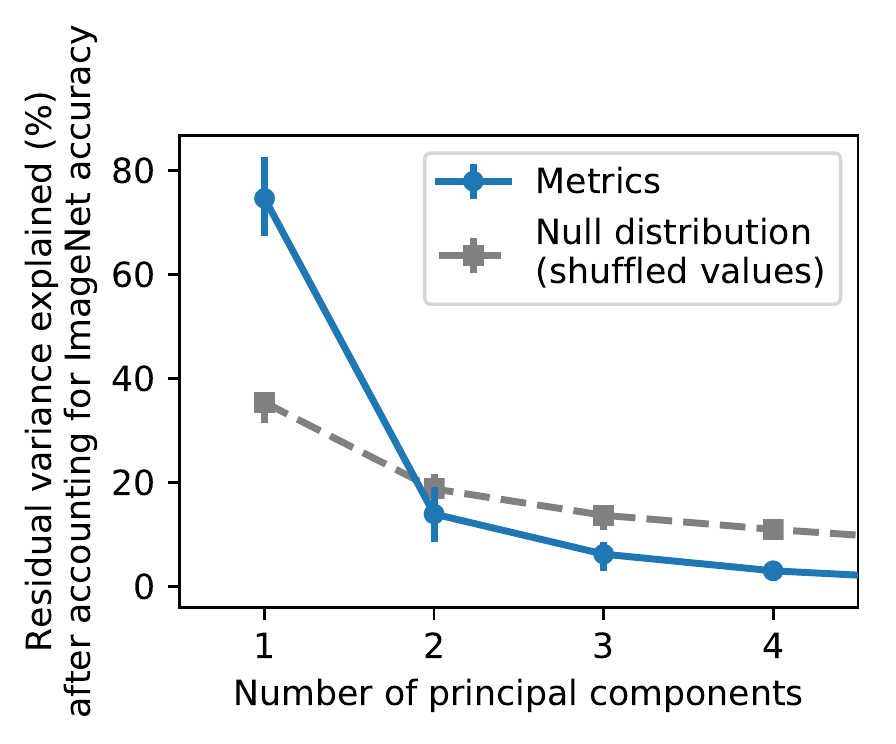}}
    \caption{\scriptsize{The space of robustness metrics.}}
    \label{fig:dimensionality_of_robustness_space}
    \end{subfigure}\hfill
    \begin{subfigure}[normal]{0.45\textwidth} 
    \centering
    \scriptsize{
    \vspace{3mm}
        \begin{tabular}{l l l} 
            \toprule
            \textsc{Dataset}             & \textsc{Instances}    & \textsc{Cls.} \\ \midrule
            \textsc{ImageNet}~\cite{krizhevsky2012imagenet}            & \num{50000}                           & $1000$ \\
            \textsc{ImageNet-A}~\cite{hendrycks2019natural}          & \num{7500}                          & $200$  \\
            \textsc{ImageNet-C}~\cite{hendrycks2018benchmarking}          & $15 \times 4 \times$\num{50000}     & $1000$ \\
            \textsc{ObjectNet}~\cite{barbu2019objectnet}           & \num{18574}                        & $113$ \\
            \textsc{ImageNet-V2}~\cite{recht2019imagenet}         & \num{10000}                          & $1000$ \\
            \textsc{ImageNet-Vid}~\cite{imnetvid}        & \num{22179}                         & $293$ \\
            \textsc{YTBB-Robust}~\cite{imnetvid}         & \num{51826}                         & $229$ \\\bottomrule
             \end{tabular}\vspace{7mm}
      \caption{\label{tab:datasets_sizes} \scriptsize{The name and reference, number of instances, and the number of classes overlapping with ImageNet for each dataset.}}
      }
    \end{subfigure}
 \caption{\textbf{(Left)} The space of robustness metrics spans approximately one statistically significant dimension after accounting for \textsc{ImageNet} accuracy. Errorbars show 95\% confidence intervals based on 1000 bootstrap samples (for the true data) or 1000 random permutations (for the null distribution). See Section~\ref{app:dimensionality_details} for details. \textbf{(Right)} Details for the datasets used in this study. The datasets were used only for evaluation.}
\end{figure*}

To estimate how many different dimensions are measured by the robustness metrics beyond what is already explained by \textsc{ImageNet} accuracy, we proceeded as follows. For each of the robustness metrics shown in Figure~\ref{fig:metric_correlations} and~\ref{fig:metric_correlations_full}, a linear regression was fit to predict that metric's value for the 39 models, using \textsc{ImageNet} accuracy as the sole predictor variable. Then, the residuals were computed for each metric by subtracting the linear regression prediction. The plot shows the fraction of variance explained for the first 4 principal components of the space of residuals of the robustness metrics. As a null hypothesis, we assumed that there is no correlation structure in the metric residuals. To construct corresponding null datasets, we randomly permuted the values for each metric independently, which destroys the correlation structure between metrics. Figure~\ref{fig:dimensionality_of_robustness_space} shows that only the first principal component is significantly above the value expected under the null hypothesis.

\subsection{Informativeness of robustness metrics}
\begin{figure}[t]
    \raisebox{1.7cm}{\includegraphics[width=0.45\textwidth]{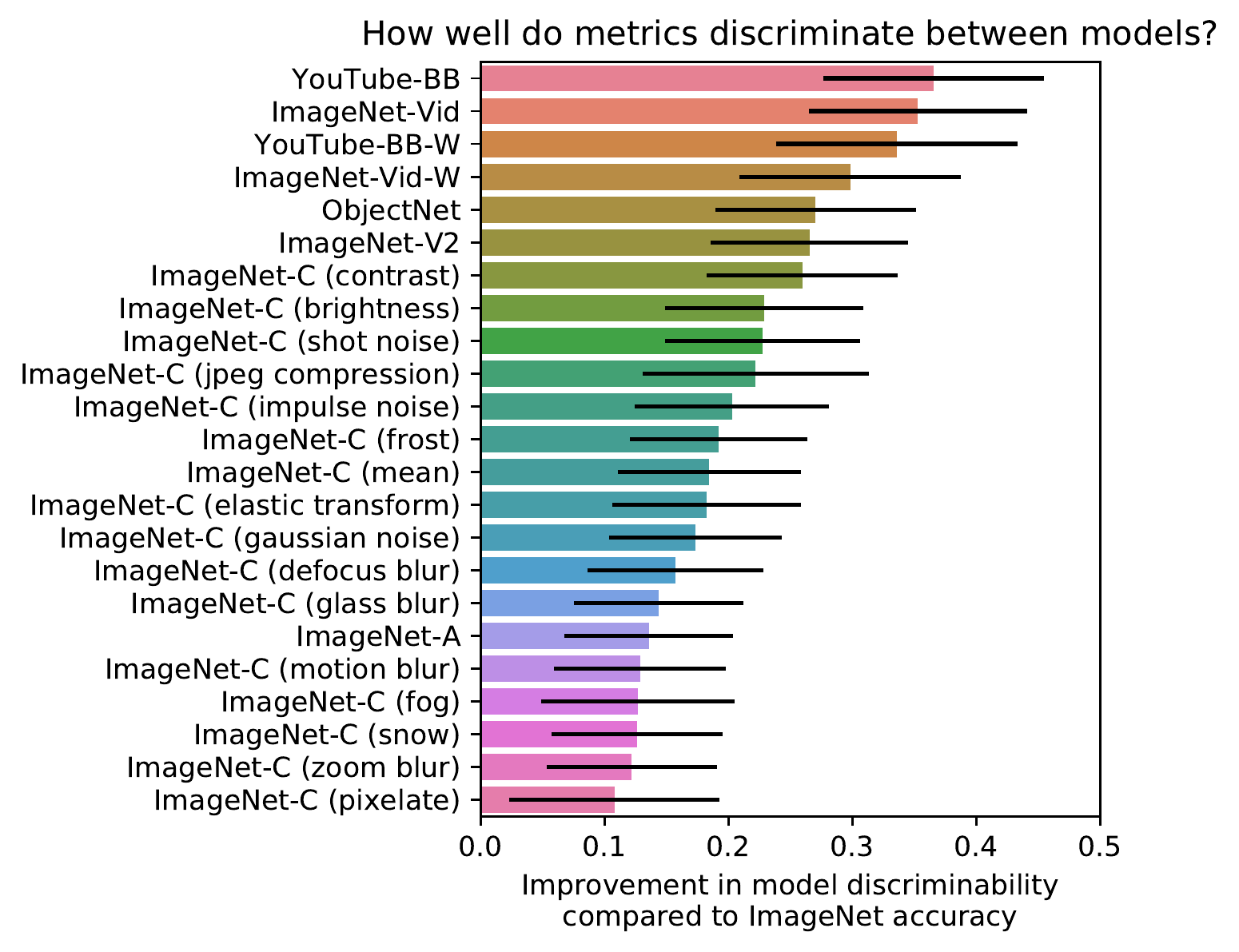}}
    \raisebox{0.0cm}{\includegraphics[width=0.54\textwidth]{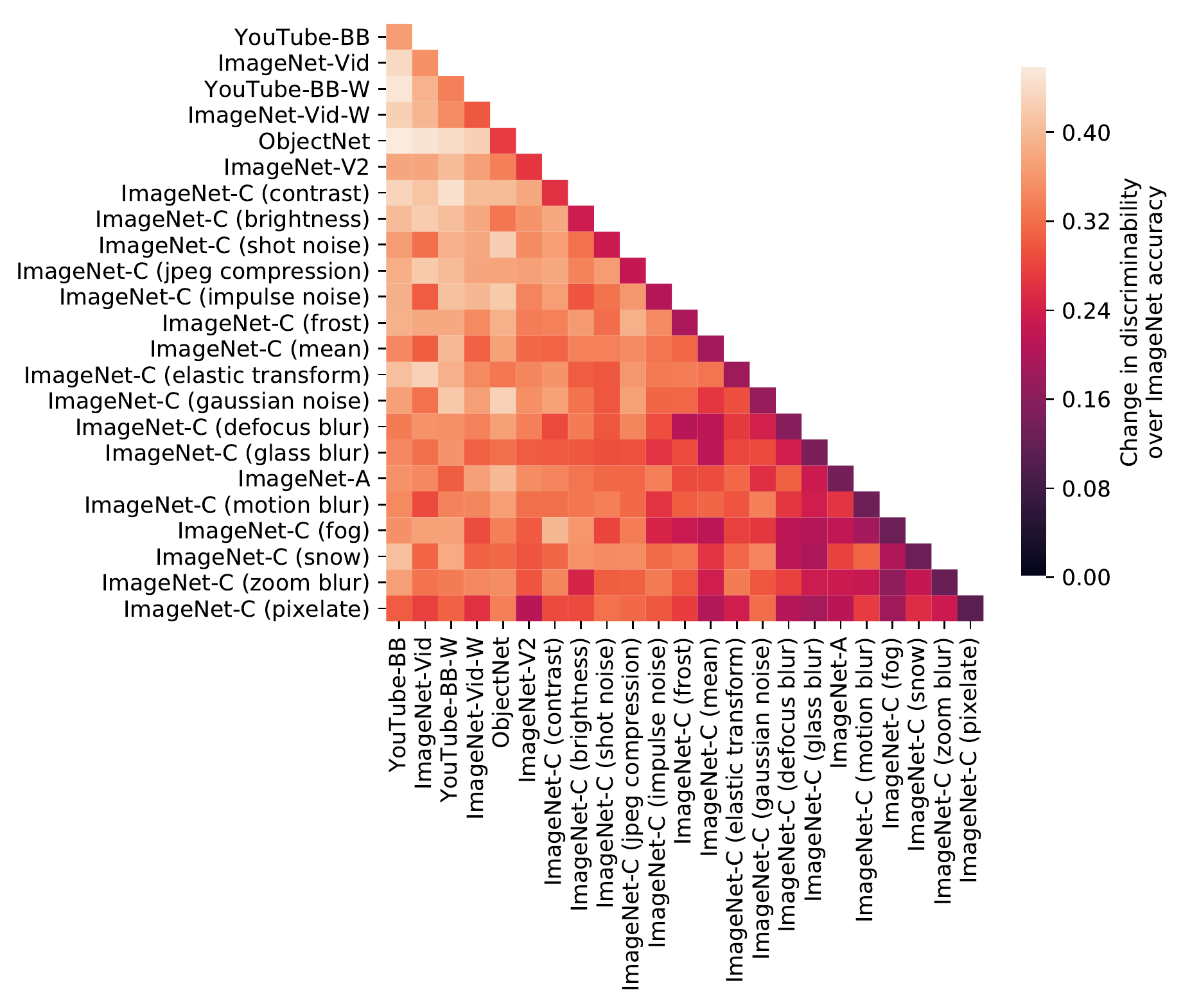}}
    \caption{\footnotesize{Informativeness of robustness metrics (related to \Cref{fig:metric_discr}).
    \textbf{(Left)} Similar to Figure~\ref{fig:metric_discr}, but showing all 23 robustness metrics. Difference in accuracy of a logistic classifier trained to discriminate between model types based on \textsc{ImageNet} accuracy plus one additional metric, compared to a classifier trained only on \textsc{ImageNet} accuracy (higher is better, top 10 metrics shown). 
    Bars show mean$\pm$s.d.\ of 1000 bootstrap samples from the 39 models.
    \textbf{(Right)} Increase in classifier accuracy over \textsc{ImageNet} accuracy when including up to two robustness metrics as explanatory variables. The diagonal shows the single-feature values from (left).}}

    \label{fig:metric_correlations_full}
\end{figure}

To estimate how useful different combinations of robustness metrics are for discriminating between model types, we trained logistic regression classifiers to discriminate between the 12 model groups outlined in the main paper. We consider \textsc{ImageNet} accuracy as a baseline metric and therefore compare the performance of a classifier using only \textsc{ImageNet} accuracy as input feature, to a classifier using \textsc{ImageNet} either one (Figure~\ref{fig:metric_correlations_full}, left) or two (Figure~\ref{fig:metric_correlations_full}, right) additional metrics as input features. Figure~\ref{fig:metric_correlations_full} shows difference in accuracy to the baseline (\textsc{ImageNet}) classifier. These results can serve practitioners with a limited budget as a rough guideline for which metric combinations are the most informative. In our experiments, the most informative combination of metrics in addition to \textsc{ImageNet} accuracy was \textsc{ObjectNet} and \textsc{YouTube-BB}, although other combinations performed similarly within the statistical uncertainty.

\subsection{Visual Task Adaptation Benchmark Details}

The Visual Task Adaptation Benchmark (VTAB)~\cite{zhai2019largescale} contains 19 tasks.
Either the full dataset or $1000$-example training sets may be used. We use the version with $1000$-example training sets (VTAB-1k).

The tasks are divided into three groups: \textit{natural} consists of standard natural image classification problems;
\textit{specialized} consists of domain-specific images captured with specialist equipment (e.g. medical images); 
\textit{structured} consists of classification tasks that require geometric understanding of a scene.
The \textit{natural} group contains the following datasets:
Caltech101 \cite{li2006one}, CIFAR-100 \cite{cifar10}, DTD \cite{cimpoi14describing}, Flowers102 \cite{Nilsback08}, Pets \cite{parkhi12a}, Sun397 \cite{xiao2010sun}, SVHN \cite{netzer2011reading}.
The \textit{specialized} group contains remote sensing datasets EuroSAT \cite{helber2017eurosat} and Resisc45 \cite{cheng2017remote},
and medical image datasets Patch Camelyon \cite{veeling2018rotation} and Diabetic Retinopathy \cite{kaggle-diabetic-retinopathy}.
The \textit{structured} group contains the following tasks:
counting and distance prediction on CLEVR \cite{johnson2017clevr},
pixel-location and orientation prediction on dSprites \cite{dsprites17}, camera elevation and object orientation on SmallNORB \cite{lecun2004learning}, object distance on DMLab \cite{beattie2016deepmind} and vehicle distance on KITTI \cite{Geiger2013IJRR}.

\section{Scale and OOD generalization}\label{app:large-scale}
\paragraph{Training Details}
The models are firstly pre-trained on \textsc{ImageNet-21k} and \textsc{JFT}, and are then fine-tuned on \textsc{ImageNet} to match the label space for evaluation. We follow the pre-training and BiT-HyperRule fine-tuning setup proposed in ~\cite{bit}.

Specifically, for pre-training, we use SGD with momentum with initial learning rate of 0.1, and momentum 0.9. 
We use linear learning rate warm-up for 5000 optimization steps and multiply the learning rate by $\frac{\text{batch size}}{256}$.
We use a weight decay of 0.0001.
We use the random image cropping technique from~\cite{inception}, and random horizontal mirroring followed by resizing the image to $224 \times 224$ pixels.
We use a global batch size of 1024 and train on a Cloud TPUv3-128.
We pre-train models for the cross product of the following combinations:
\begin{itemize}
  \item \textbf{Dataset Size}: \{1.28M (1$\times$ ImageNet train set), 2.6M (2$\times$ ImageNet train set), 5.2M (4$\times$ ImageNet train set), 9M (7$\times$ ImageNet train set), 13M (10$\times$ ImageNet train set)\}.
  \item \textbf{Train Schedule} (steps): \{113K (90 ImageNet epochs), 229K (180 ImageNet epochs), 457K (360 ImageNet epochs), 791K (630 ImageNet epochs), 1.1M (900 ImageNet epochs)\}.
\end{itemize}

For fine-tuning, we use the BiT-Hyperrule as described in ~\cite{bit}: batch size 512, learning rate 0.003, no weight decay, the classification head initialized to zeros, Mixup \cite{DBLP:conf/iclr/ZhangCDL18} with $\alpha=0.1$, fine-tuning for \num{20000} steps with $384 \times 384$ image resolution.

\paragraph{Additional Results}
Here we highlight the results equivalent to Figure~\ref{fig:model_size}, with the only difference that we consider subsets of the JFT~\cite{DBLP:conf/iccv/SunSSG17} dataset, instead of \textsc{ImageNet-21k} (Figure~\ref{fig:model_size_jft}). 
We present the results on the synthetic dataset in Appendix~\ref{app:synth}.
\begin{figure*}[h!]
    \centering
    \includegraphics[width=\linewidth]{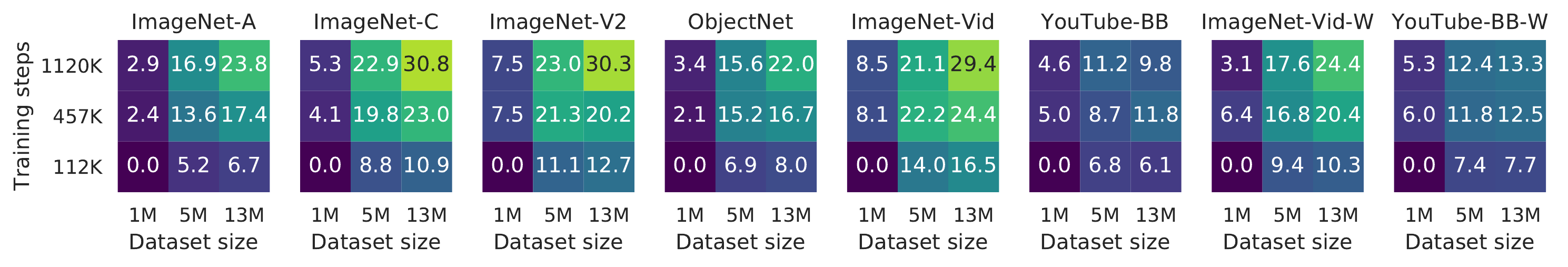}
    \includegraphics[width=\linewidth]{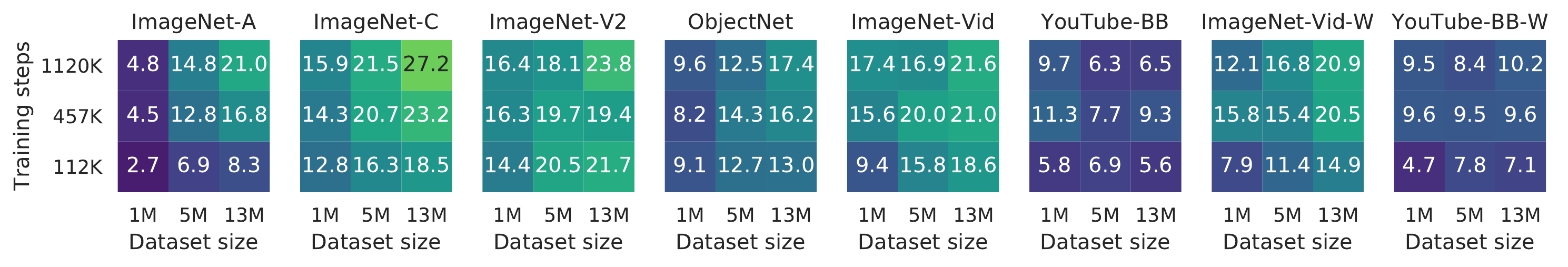}
    \vspace{-5mm}
    \caption{\textbf{(Top)}  Reduction (in \%) in classification error relative to the classification error of the model trained for 112k steps on 1M examples (bottom left corner) as a function of training steps and training set size. The results are for ResNet-50 trained on \textsc{JFT} subsets. \textbf{(Bottom)} Relative reduction (in \%) in classification error going from ResNet-50 to ResNet-101x3 as a function of training steps and training set size (\textsc{JFT} subsets). The reduction generally increases with the training set size and longer training.}
    \label{fig:model_size_jft}
\end{figure*}

\section{Effect of the testing resolution}\label{app:resolution}

\paragraph{Cropping details} Before applying the respective model, we first resize every image such that the shorter side has length $\lfloor 1.15 \cdot r \rfloor$ while preserving the aspect ratio and take a central crop of size $r \times r$. For the widely used $224 \times 224$ testing resolution, this leads to standard single-crop testing preprocessing, where the images are first resized such that the shorter side has length $256$.

\paragraph{Training details for FixRes} For fine-tuning to the target resolution (FixRes) we use SGD with momentum with initial learning rate of $0.004$ (except for the BiT models for which we use $0.0004$), and momentum 0.9, accounting for varying batch size by multiplying the learning rate with $\frac{\text{batch size}}{256}$. We train for \num{15000}$\cdot \frac{\text{batch size}}{2048}$, decaying the learning rate by a factor of $10$ after $1/3$ and $2/3$ of the iterations. The batch size is chosen based on the model size to avoid memory overflow; we use $2048$ in most cases. We train on a Cloud TPUv3-64. We emphasize that we did not extensively tune the training parameters for FixRes, but chose a setting that works well across models and data sets.

\begin{figure}[ht]
    \centering
    \begin{subfigure}[normal]{\textwidth} 
        \includegraphics[width=\textwidth]{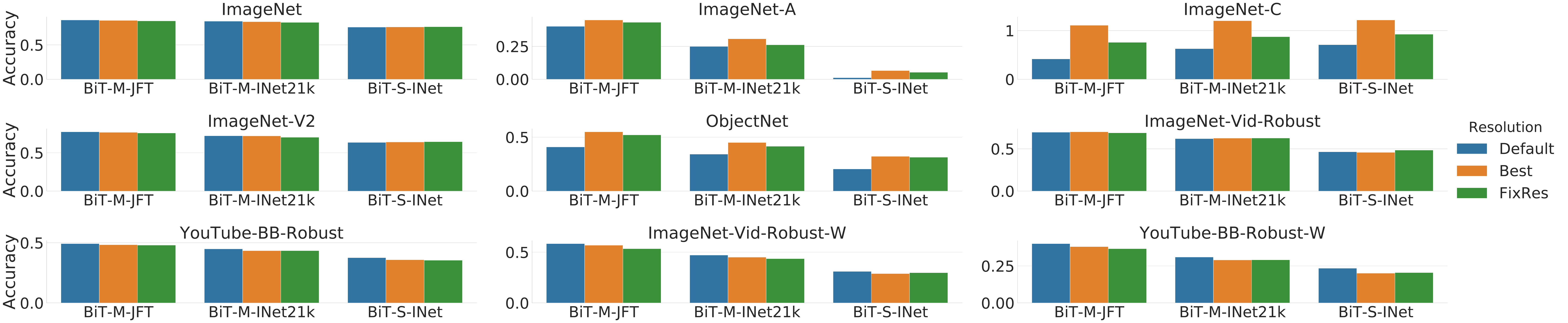}
        \caption{Different BiT variants. -M- stands for ResNet-101x3, while -S- stands for ResNet-50x1. INet is a shorthand for ImageNet.}
    \end{subfigure}
    \begin{subfigure}[normal]{\textwidth} 
        \includegraphics[width=\textwidth]{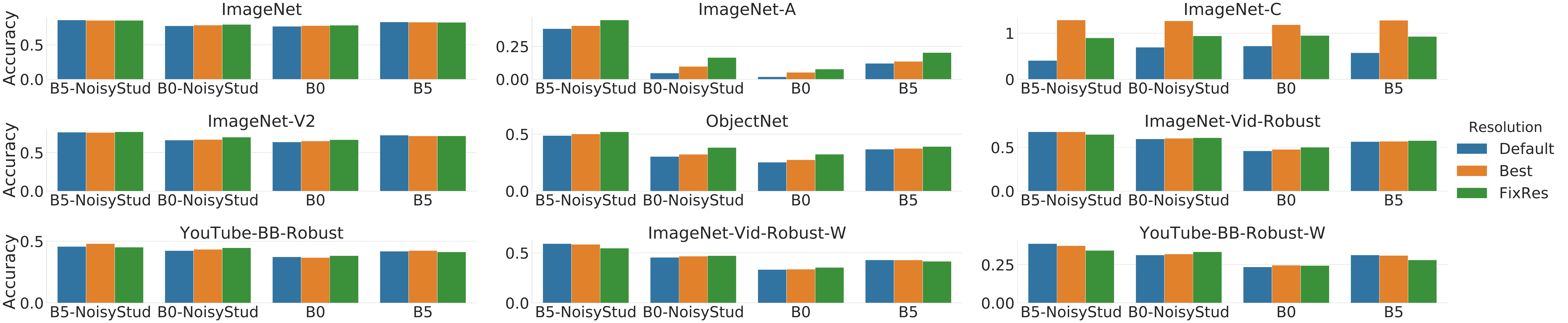}
        \caption{Two ImageNet-trained EfficientNet variants (B0,B5) as well as those models trained using the Noisy Student protocol.}      
    \end{subfigure}
    \begin{subfigure}[normal]{\textwidth} 
      \includegraphics[width=\textwidth]{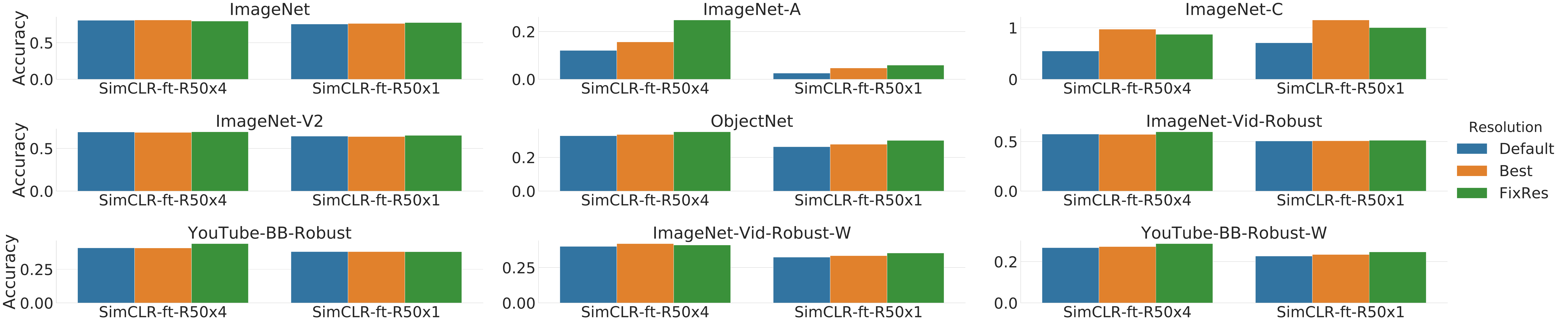}
        \caption{SimCLR models that have been fine-tuned on ImageNet.}
    \end{subfigure}
    \begin{subfigure}[normal]{\textwidth} 
        \includegraphics[width=\textwidth]{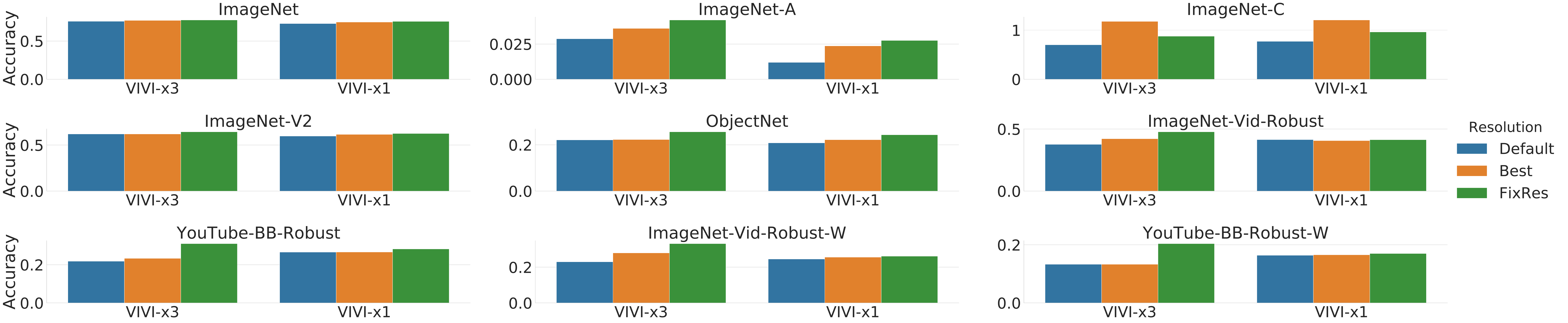}
        \caption{Two VIVI variants (R50x1 and R50x3), both co-trained with ImageNet.}
    \end{subfigure}
    \caption{Comparison of different types of evaluation preprocessing and resolutions. Default: Accuracy obtained for the preprocessing and resolution proposed by the authors of the respective models. Best: The accuracy when selecting the best resolution from $\{64, 128, 224, 288, 320, 384, 512, 768\}$. FixRes: Applying FixRes for the same set of resolutions and selecting the best resolution. Increasing the evaluation resolution and additionally using FixRes helps across a large range of models and pretraining datasets.}\label{fig:resolution-all-models}
\end{figure}

\begin{figure}[h]
    \centering
    \includegraphics[width=1\textwidth]{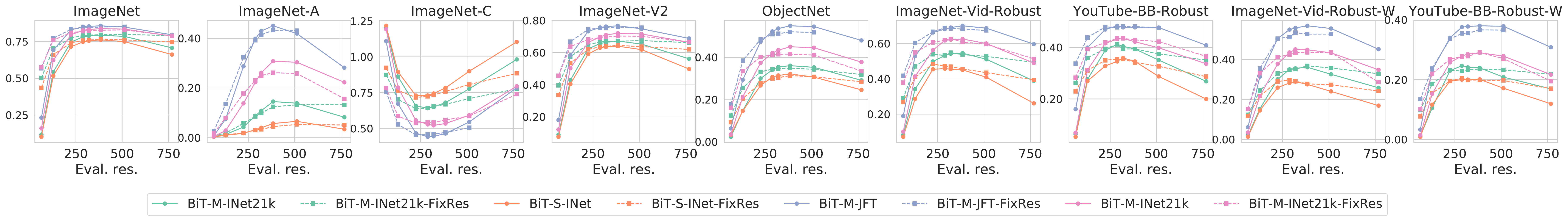}
    \includegraphics[width=1\textwidth]{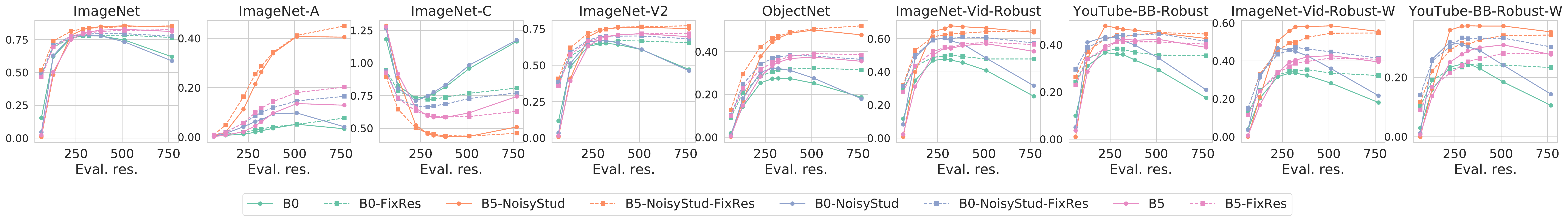}
    \includegraphics[width=1\textwidth]{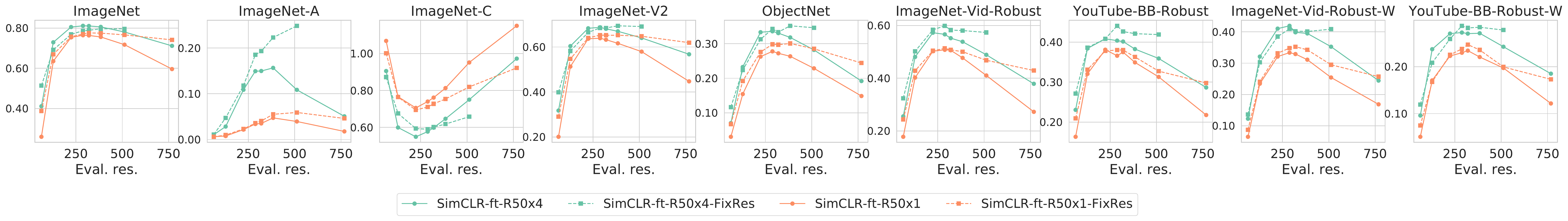}
    \includegraphics[width=1\textwidth]{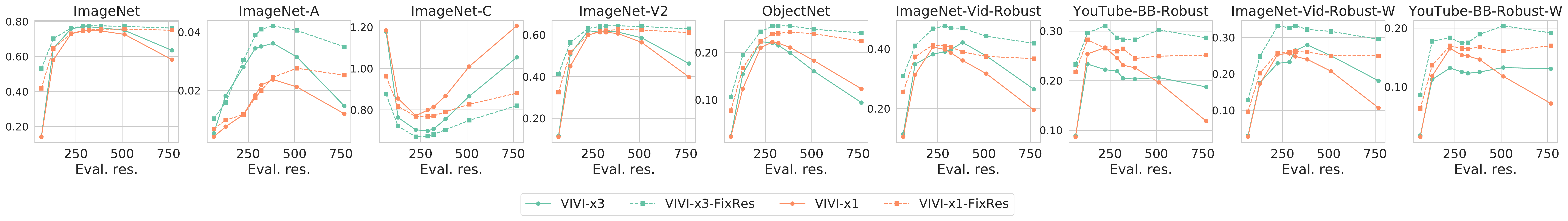}
    \includegraphics[width=1\textwidth]{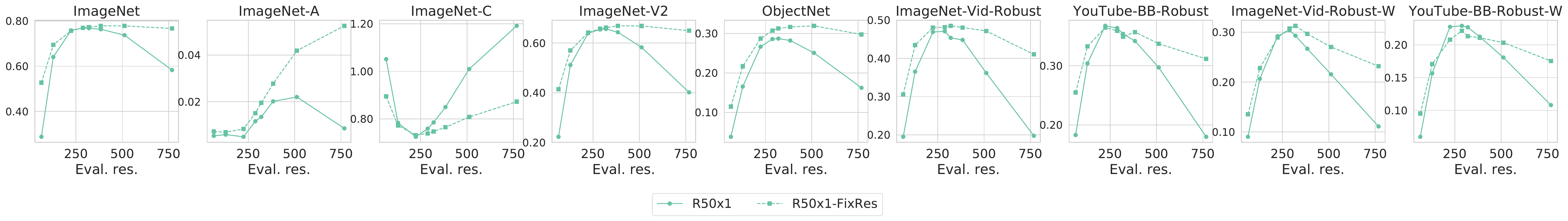}
    \caption{Comparison of different types of evaluation preprocessing and resolutions, without modifying the model and after applying FixRes. For brevity the same shorthands are used in the model names as in \Cref{fig:resolution-all-models}.}
    \label{fig:fixres-full}
\end{figure}

\paragraph{Additional results} In \Cref{fig:resolution-all-models} we provide an extended version of \Cref{fig:fixres} that shows the effect of FixRes for all datasets and models. In \Cref{fig:fixres-full} we plot the performance of all models and their FixRes variants as a function of the resolution.

\newpage
\newpage

\section{Additional results on \textsc{SI-Score}, the synthetic dataset}\label{app:synth}

\vspace{-5mm}
\begin{figure}[b]
\begin{subfigure}[b]{0.49\textwidth}
    \centering
    \includegraphics[width=0.175\linewidth]{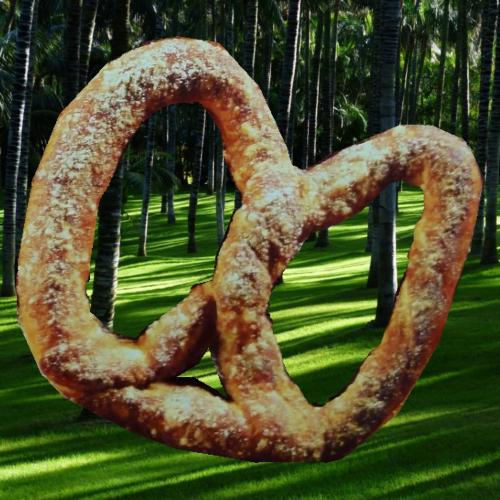}
    \includegraphics[width=0.175\linewidth]{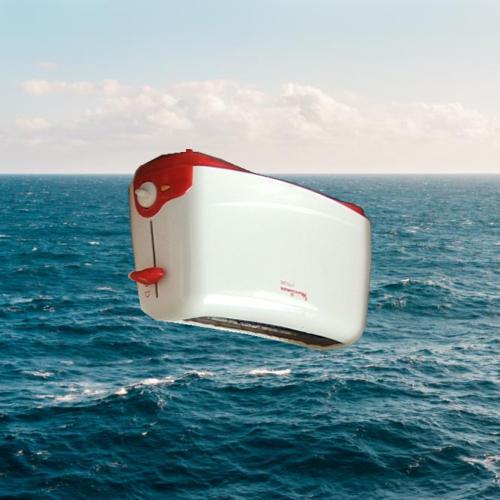}
    \includegraphics[width=0.175\linewidth]{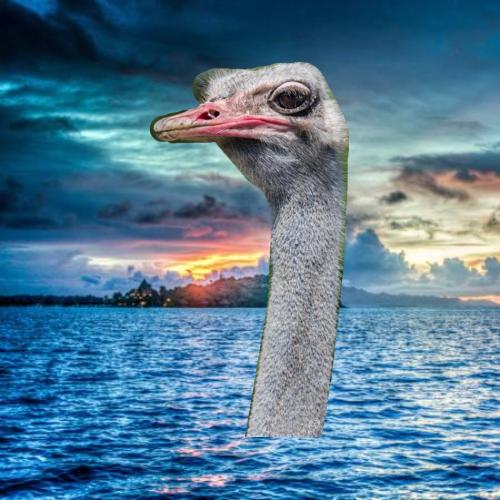}
    \includegraphics[width=0.175\linewidth]{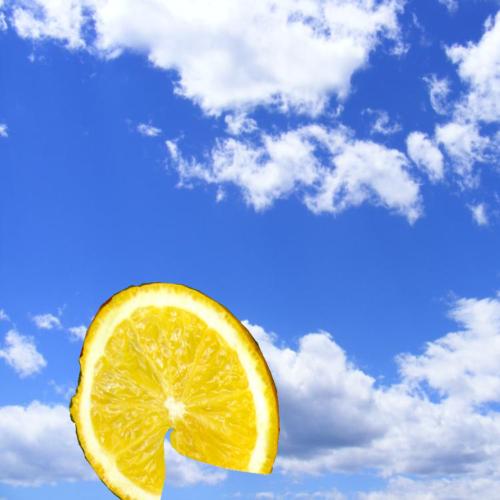} \\
    \includegraphics[width=0.175\linewidth]{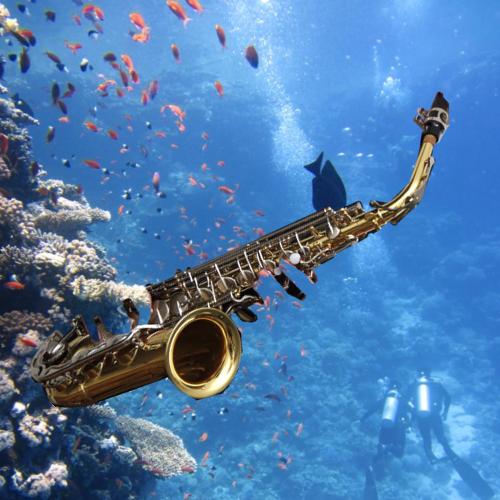}
    \includegraphics[width=0.175\linewidth]{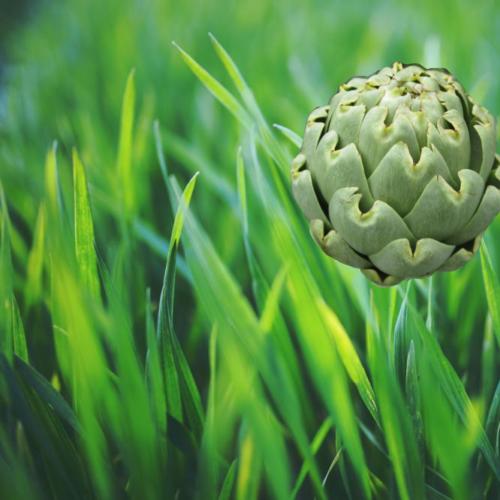}
    \includegraphics[width=0.175\linewidth]{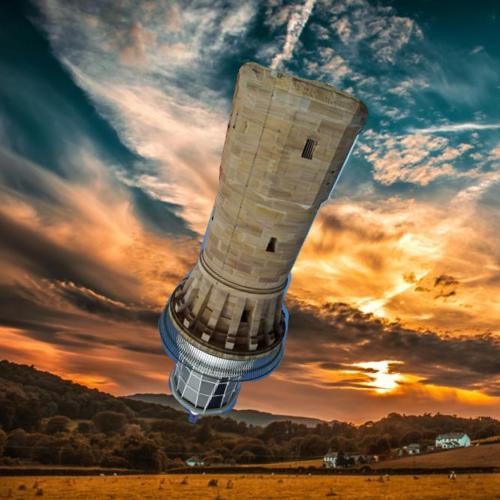}
    \includegraphics[width=0.175\linewidth]{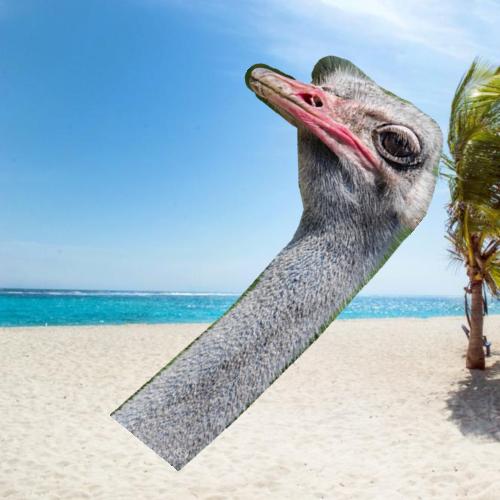} \\
    \includegraphics[width=0.175\linewidth]{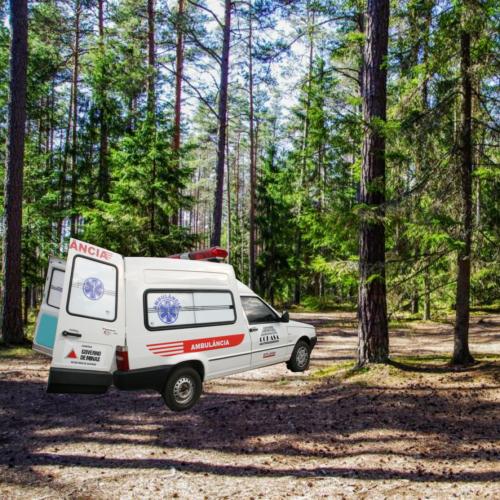}
    \includegraphics[width=0.175\linewidth]{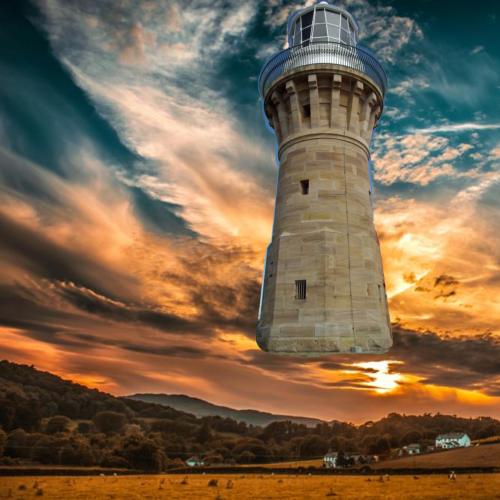}
    \includegraphics[width=0.175\linewidth]{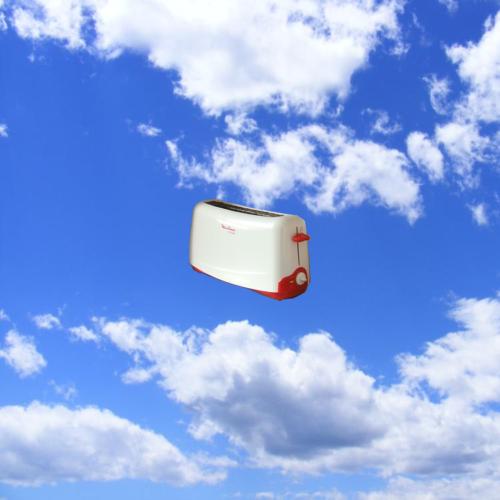}
    \includegraphics[width=0.175\linewidth]{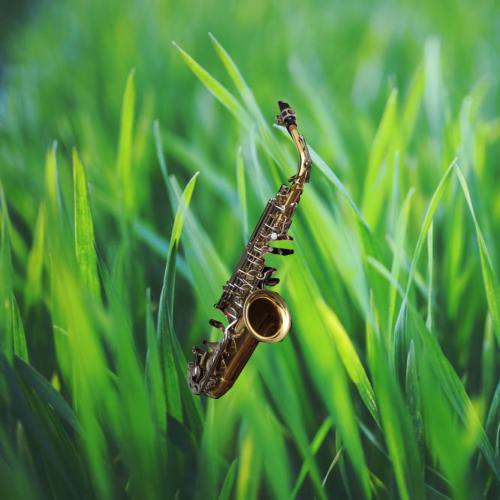} \\
        \caption{}
        \label{fig:synth-examples-appendix}
        \end{subfigure}
    \begin{subfigure}[b]{0.49\textwidth}
    \includegraphics[width=0.80\linewidth]{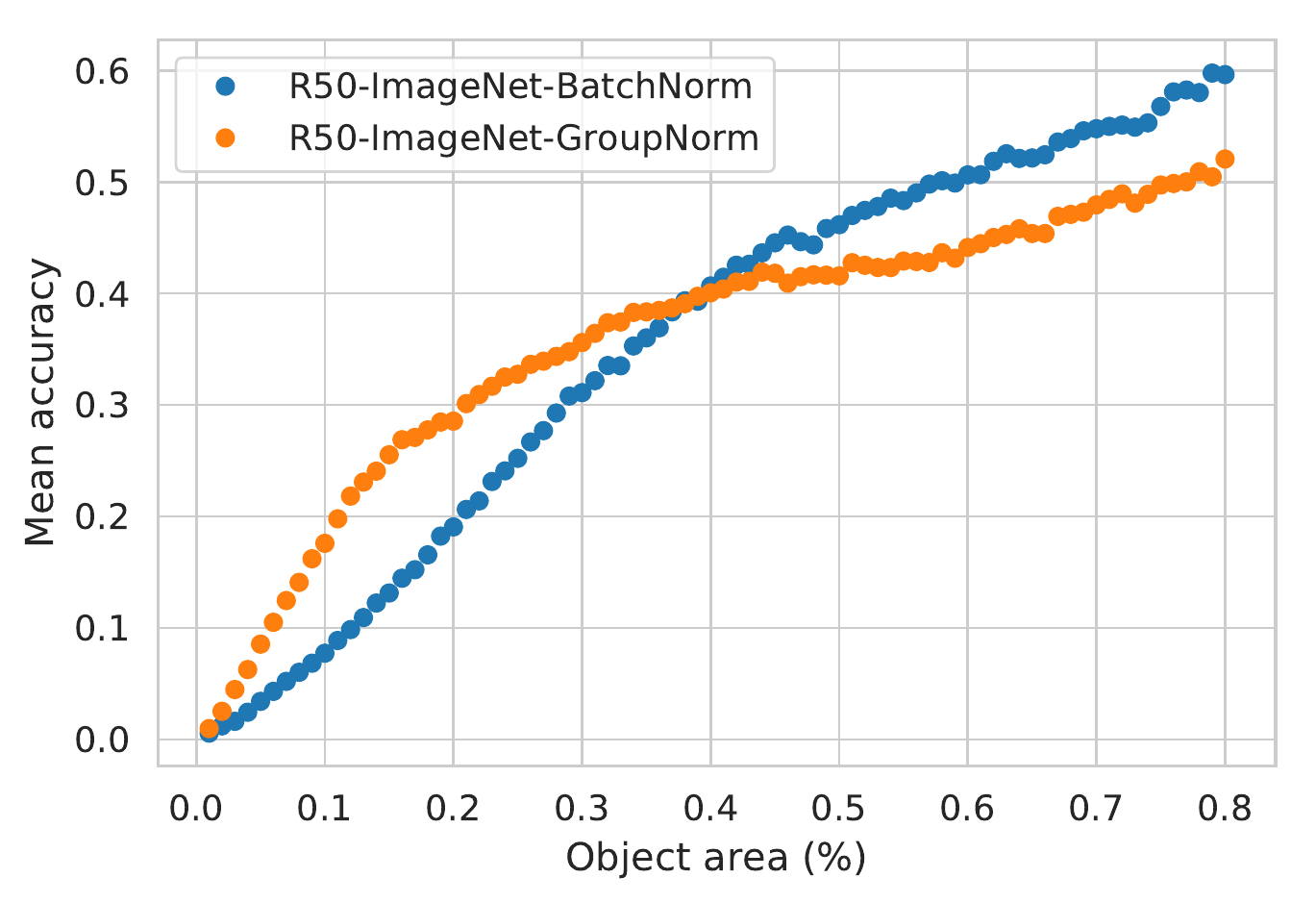}
    \caption{}
    \label{fig:synth-gn-bn}
    \end{subfigure}
    \caption{(\textbf{Left}) Additional sample images from our synthetic dataset. (\textbf{Right}) From \textsc{SI-Score}, we find that an ImageNet-trained ResNet-50 has higher classification accuracy on smaller objects if it uses GroupNorm and higher accuracy on larger objects if it uses BatchNorm. Investigating this phenomena in detail is outside the scope of this paper - here we simply highlight the potential of investigating models using datasets such as \textsc{SI-Score}.}
    \end{figure}

\begin{figure*}
    \centering
    \includegraphics[width=.65\linewidth]{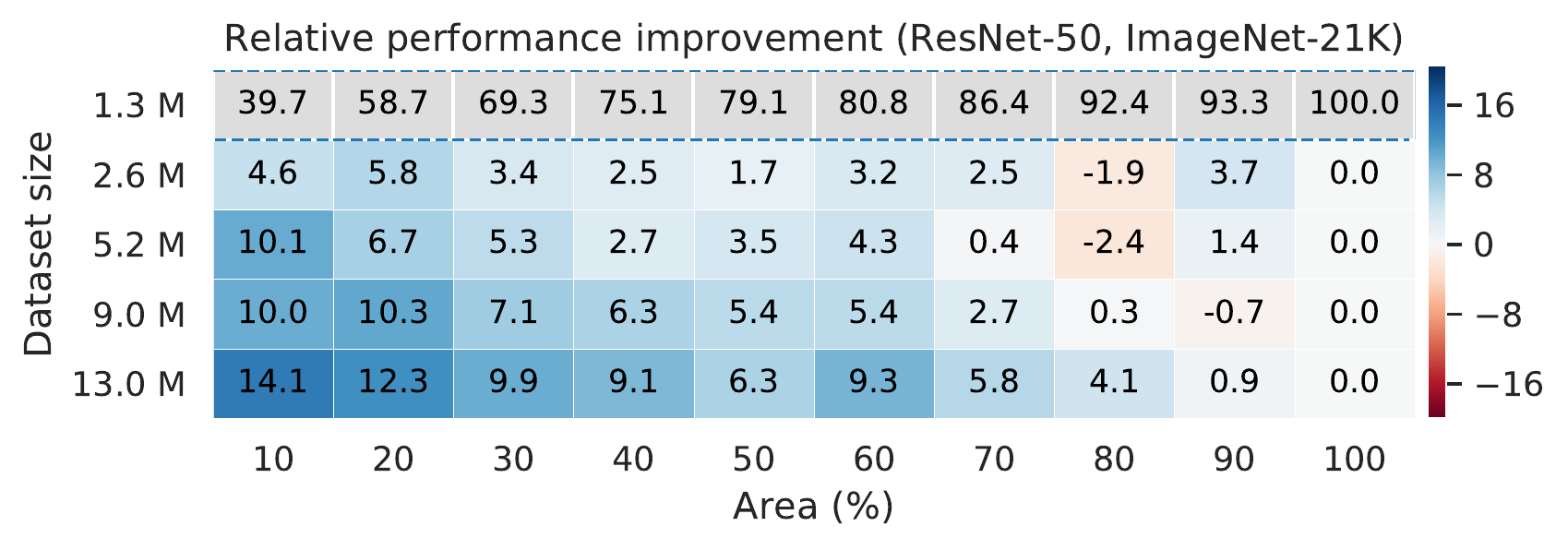}
    \includegraphics[width=.65\linewidth]{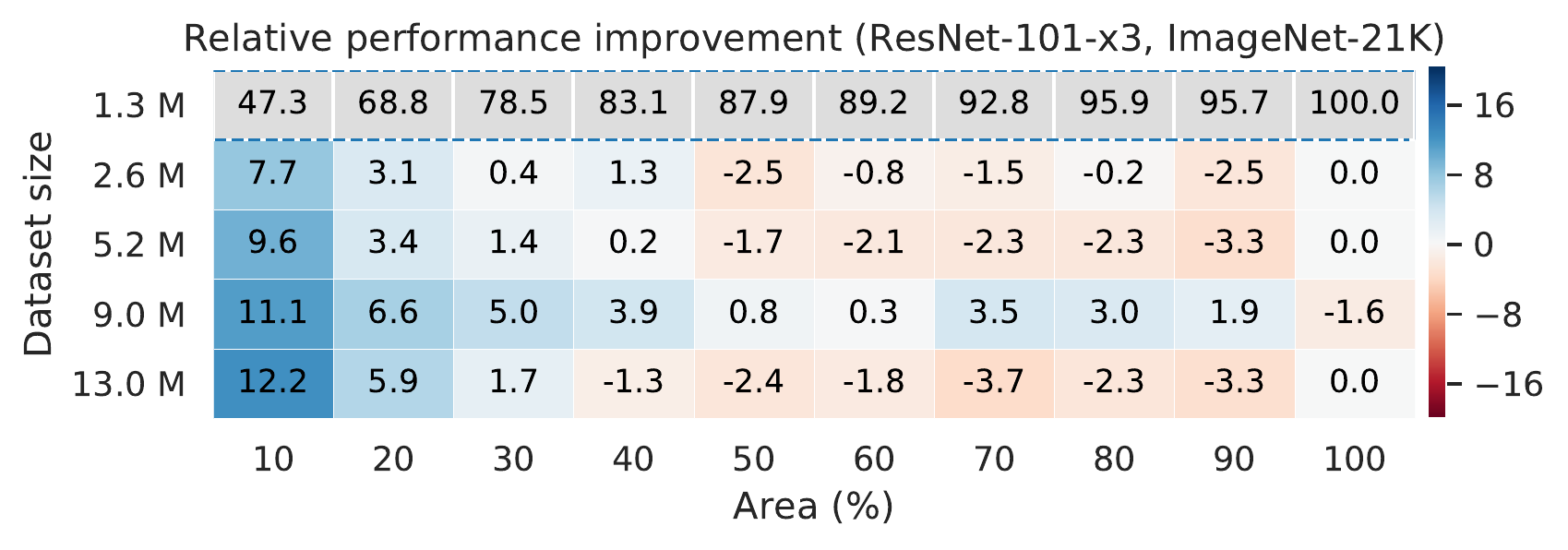}
    \includegraphics[width=.65\linewidth]{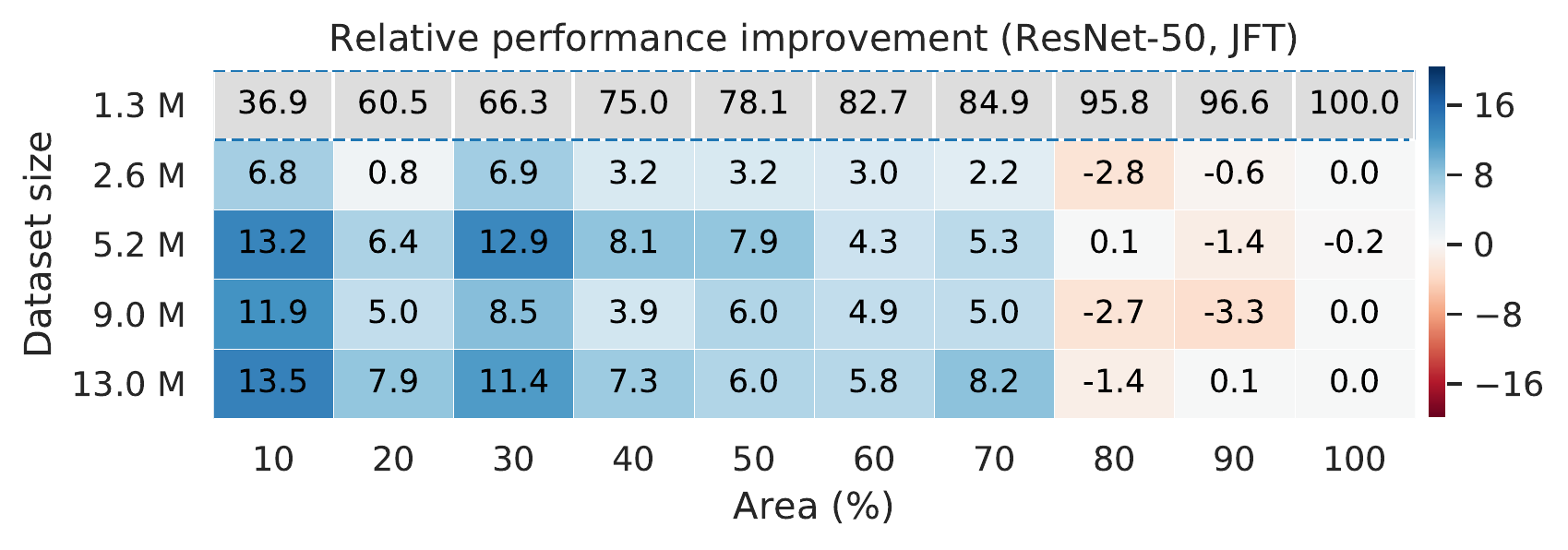}
    \includegraphics[width=.65\linewidth]{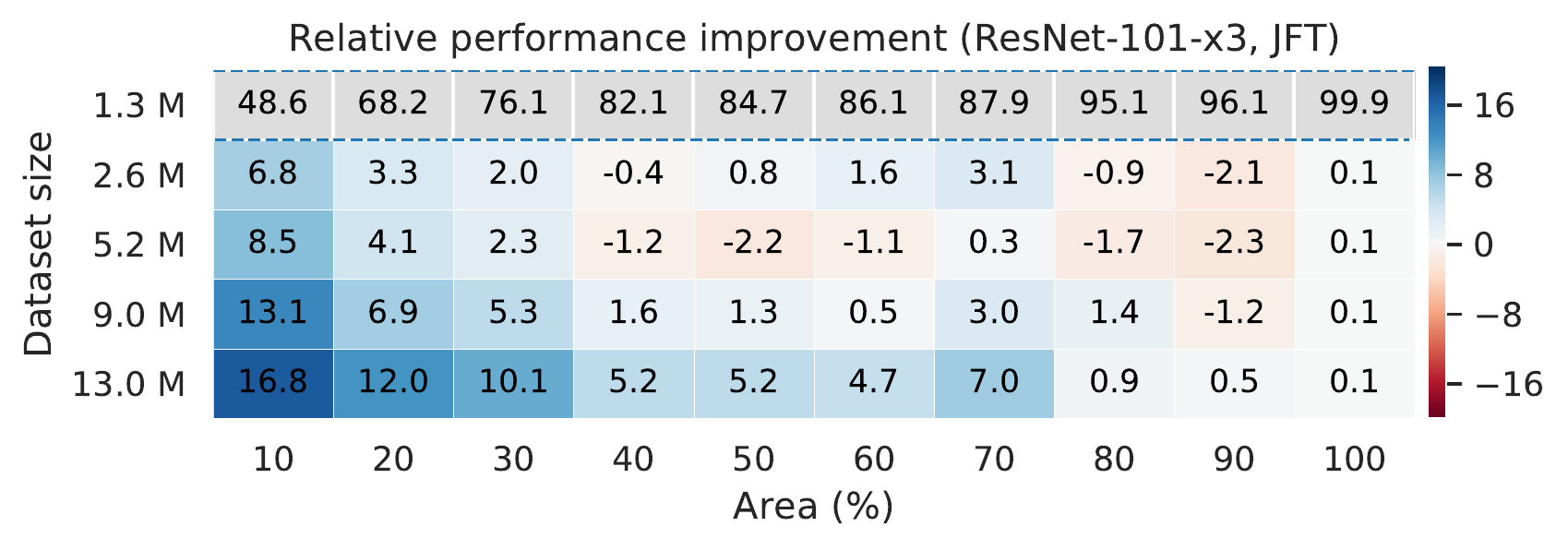}
    \caption{In the first row of both plots we show the ratio of the accuracy and the best accuracy (across all areas). For the second row (model trained on 2.6M instances),  and other rows, we compute the same normalized score and visualize the difference with the first row. Larger differences imply a more uniform behavior across relative object areas. We observe that, as the dataset size increases, the average prediction accuracy across various object areas becomes more uniform. The effect is more pronounced for the larger model. As expected, the improvement is most pronounced for small object sizes covering $10$-$20\%$ of the pixels.}
    \label{fig:synth-area}
\end{figure*}

\begin{figure*}
    \centering
    \includegraphics[width=0.95\linewidth]{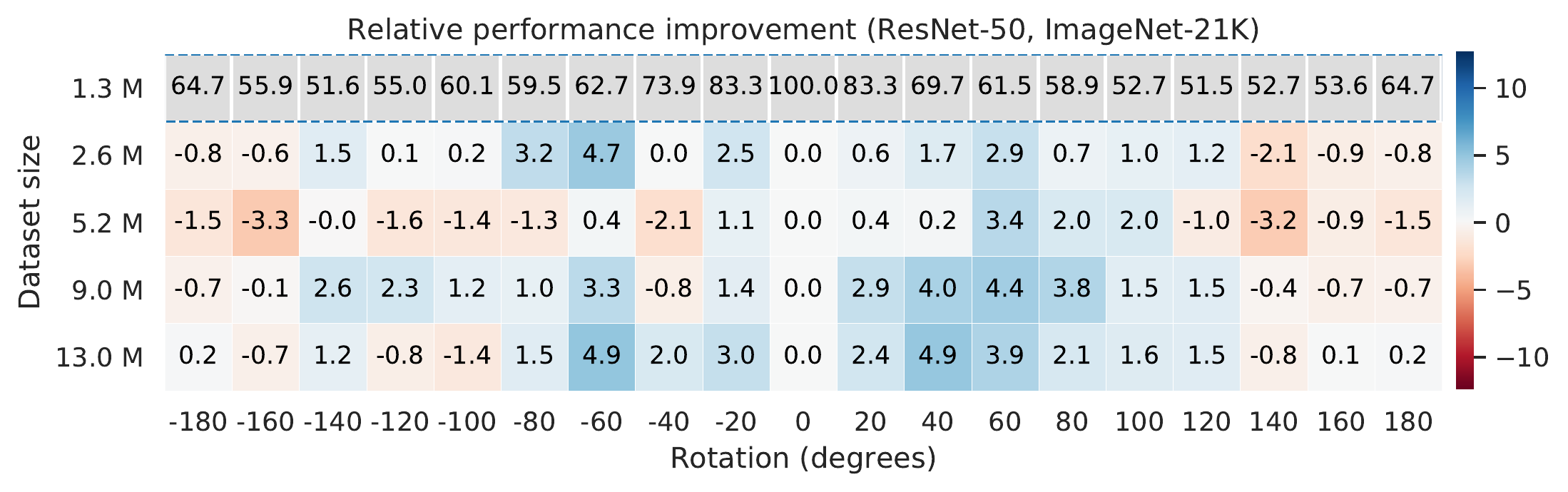}
    \includegraphics[width=0.95\linewidth]{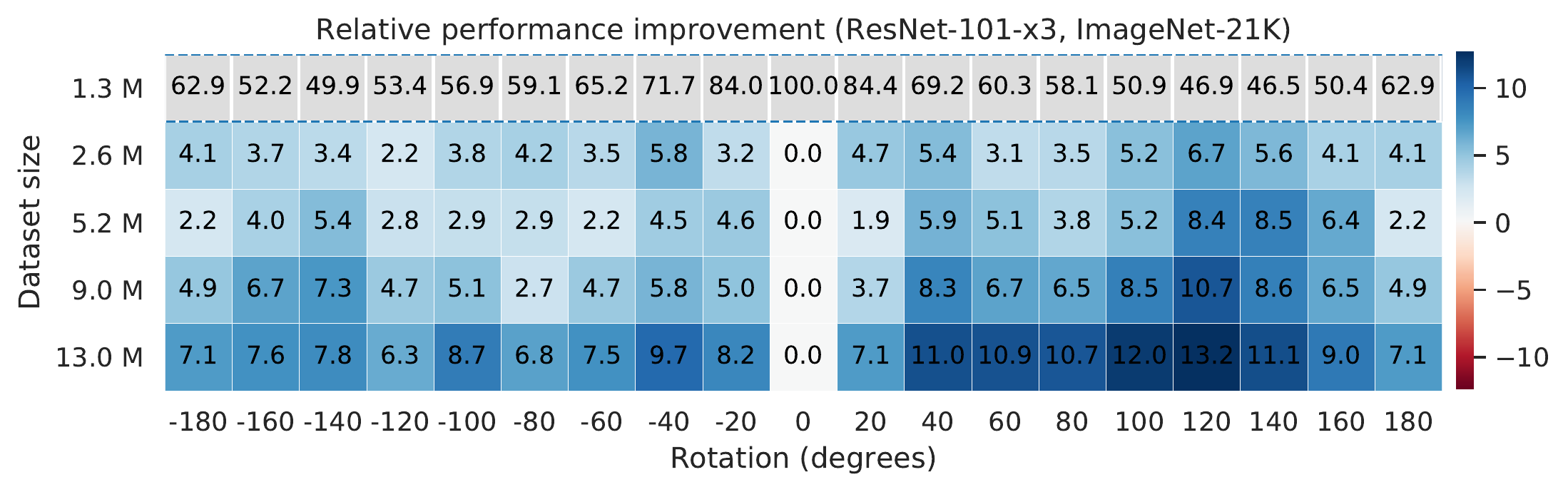}
    \includegraphics[width=0.95\linewidth]{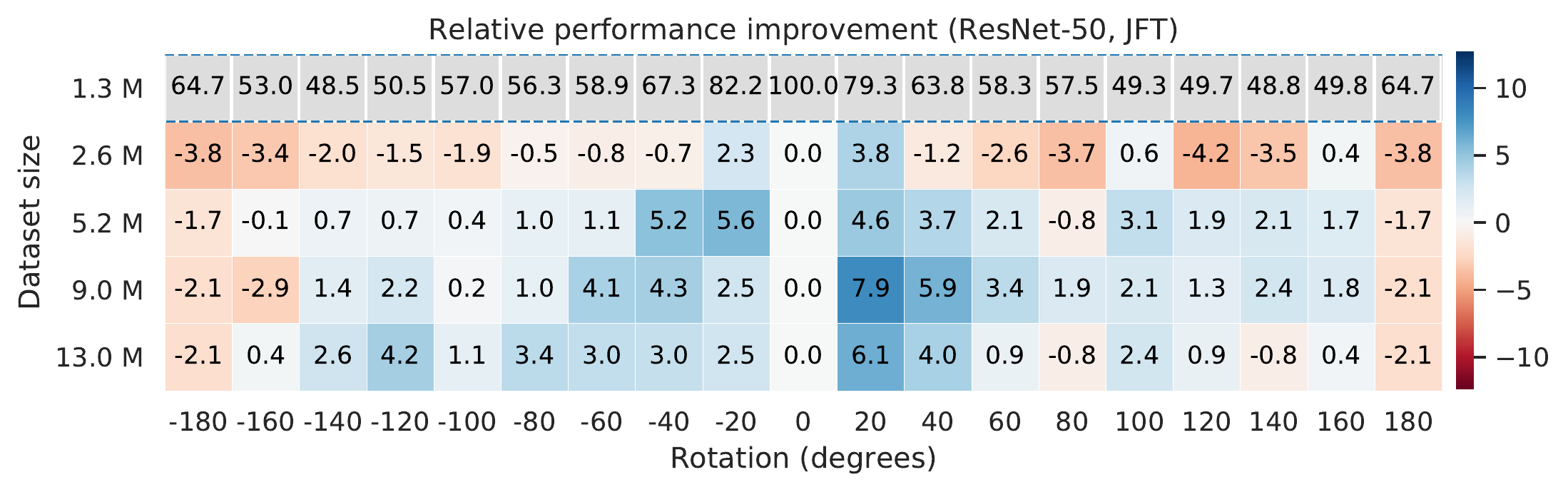}
    \includegraphics[width=0.95\linewidth]{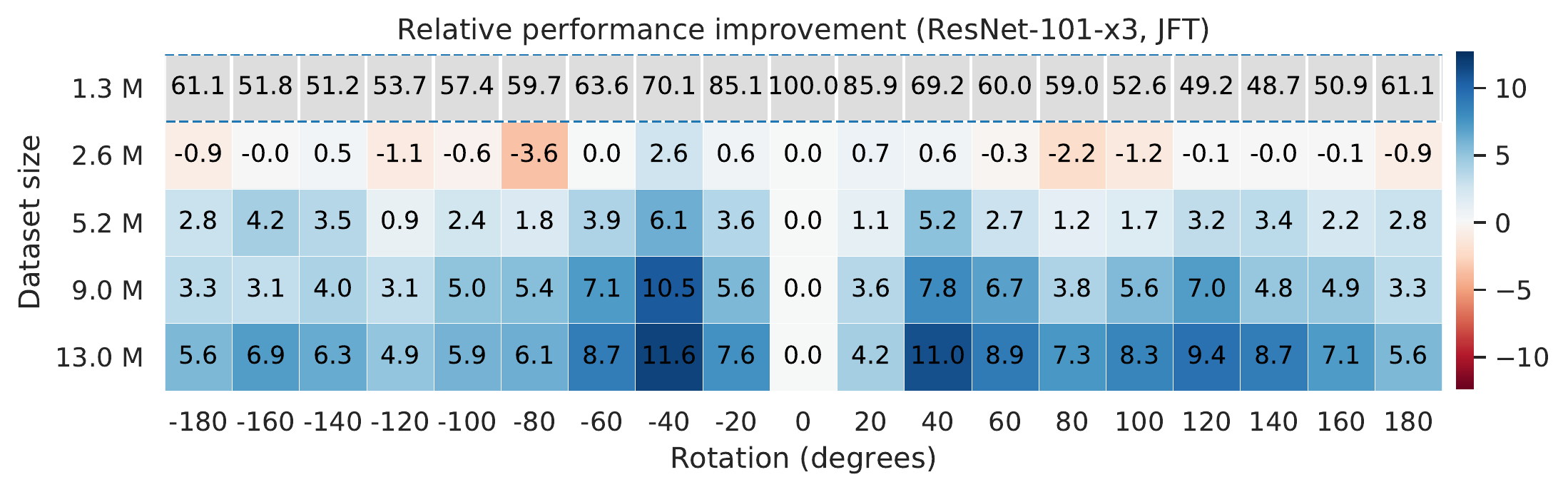}
    \caption{In the first row of both plots we show the ratio of the accuracy and the best accuracy (across all rotations). For the second row (model trained on 2.6M instances), and other rows, we compute the same normalized score and visualize the difference with the first row. Larger differences imply a more uniform behavior across object rotations. We observe that, as the dataset size increases, the average prediction accuracy across various rotation angles becomes more uniform. The effect is more pronounced for the larger model.}
    \label{fig:synth-rotation}
\end{figure*}

\begin{figure*}
    \centering
    \includegraphics[width=.9\linewidth]{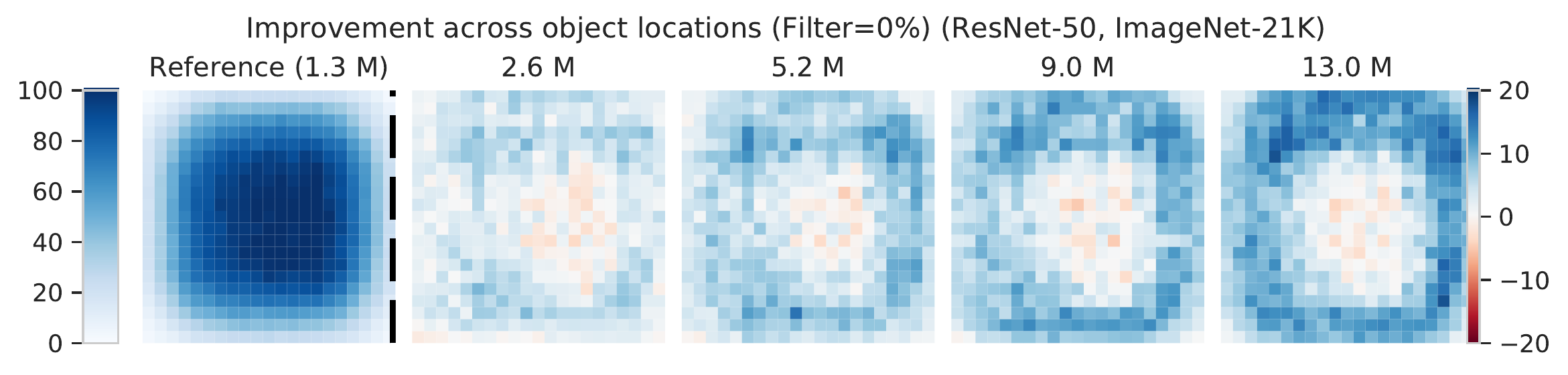}
    \includegraphics[width=.9\linewidth]{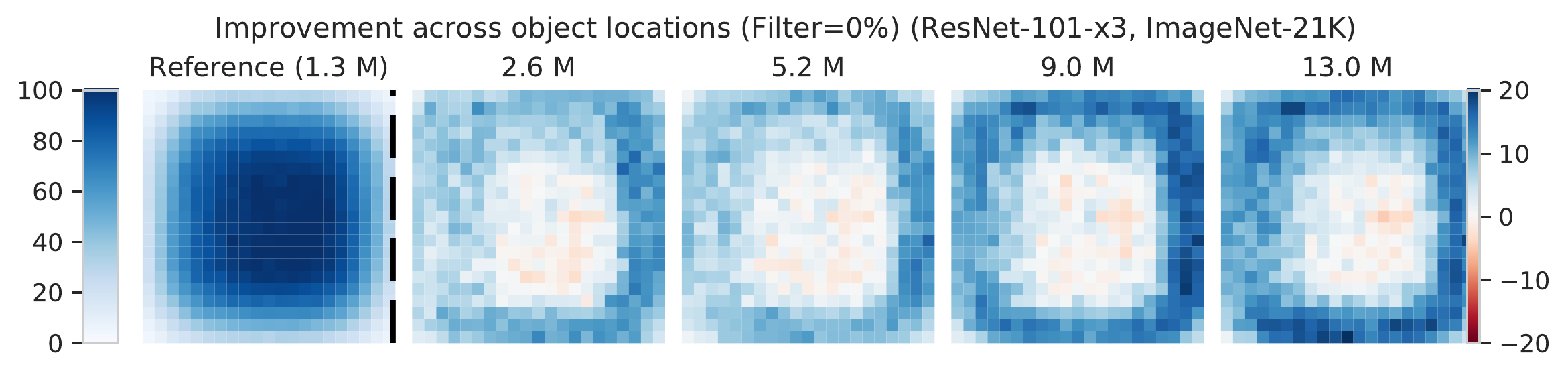}
    \includegraphics[width=.9\linewidth]{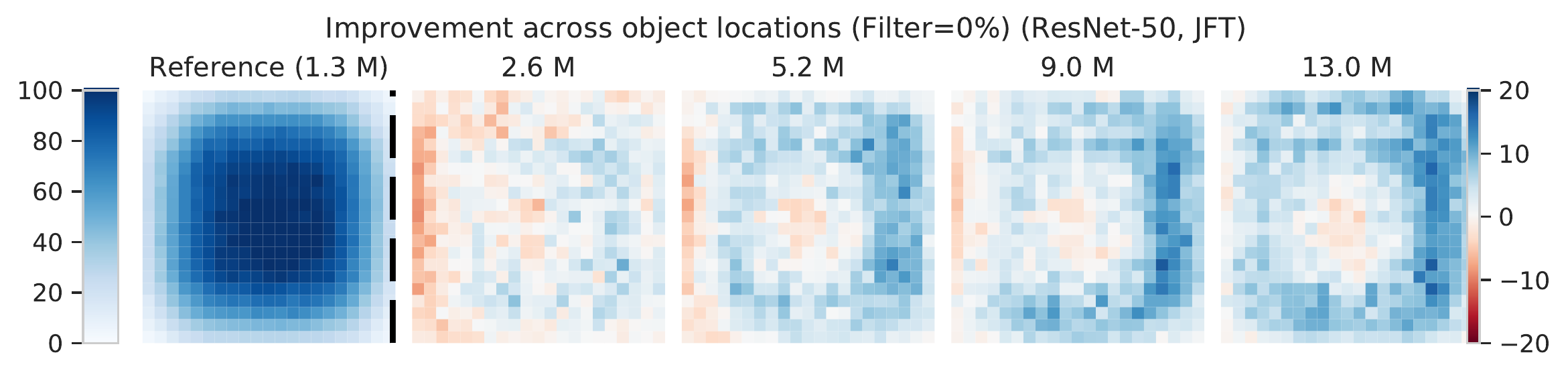}
    \includegraphics[width=.9\linewidth]{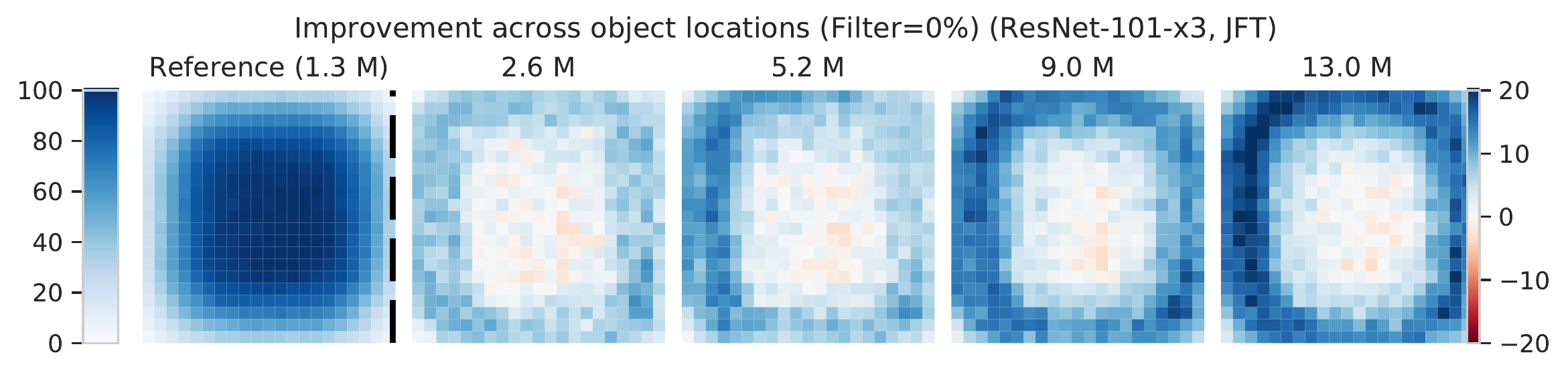}
    \caption{In the first column, for each location on the grid, we compute the average accuracy. Then, we normalize each location by the 95\textsuperscript{th} percentile across all locations, which quantifies the gap between the locations where the model performs well, and the ones where it under-performs (first column, dark blue vs white). Then, we consider models trained with more data, compute the same normalized score, and plot the \emph{difference} with respect to the first column. We observe that, as dataset size increases, sensitivity to object location decreases -- the outer regions improve in relative accuracy more than the inner ones (e.g. dark blue vs white in the second to fifth columns). The effect is more pronounced for the larger model.}
    \label{fig:synth-location-0pc}
\end{figure*}

\begin{figure*}
    \centering
    \includegraphics[width=.9\linewidth]{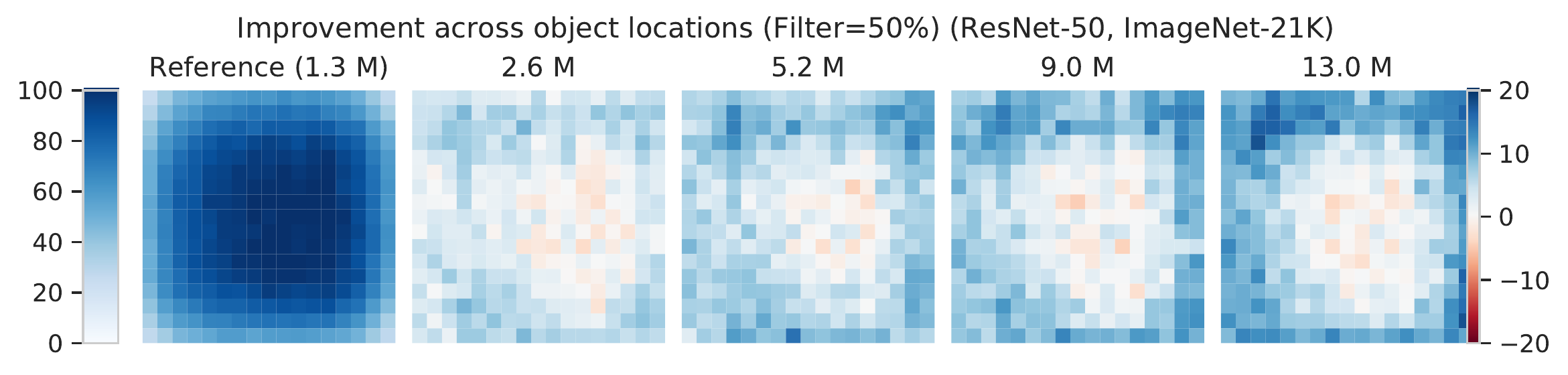}
    \includegraphics[width=.9\linewidth]{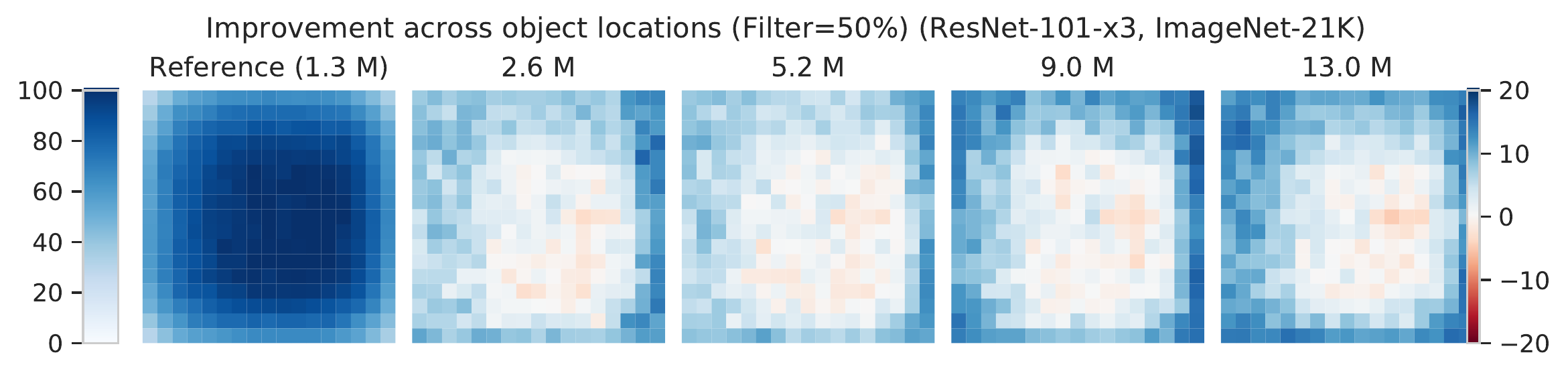}
    \includegraphics[width=.9\linewidth]{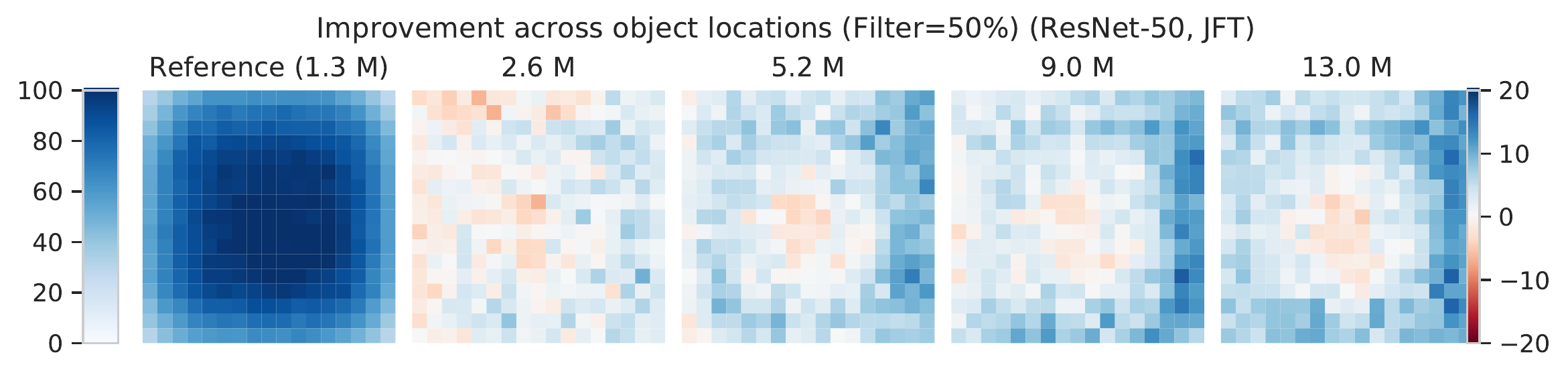}
    \includegraphics[width=.9\linewidth]{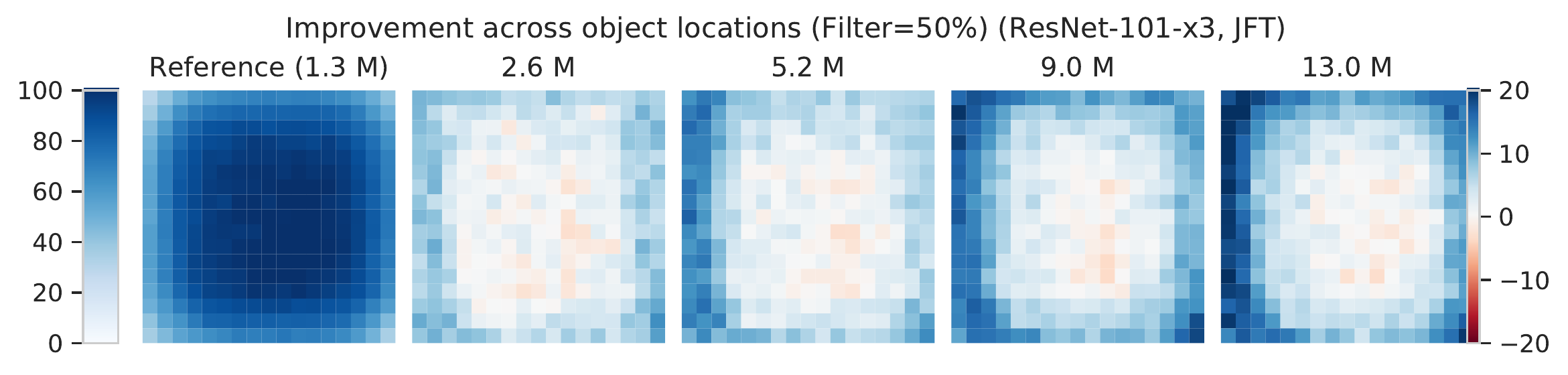}
    \caption{In the main paper, we presented results on the location dataset when not filtering out images where the objects were partially occluded, since that would exclude many locations from the dataset. For completeness, we present results filtering out objects that are less than 50\% or 75\% in the image in this figure and Figure~\ref{fig:synth-location-75pc}. \newline In the first column, for each location on the grid, we compute the average accuracy. Then, we normalize each location by the 95\textsuperscript{th} percentile across all locations, which quantifies the gap between the locations where the model performs well, and the ones where it under-performs (first column, dark blue vs white). Then, we consider models trained with more data, compute the same normalized score, and plot the \emph{difference} with respect to the first column. We observe that, as dataset size increases, sensitivity to object location decreases -- the outer regions improve in relative accuracy slightly more than the inner ones (e.g. dark blue vs white in the second to fifth columns). The effect is more pronounced for the larger model. We filter out all test images for which the foreground object is not at least $50\%$ within the image.}
    \label{fig:synth-location-50pc}
\end{figure*}

\begin{figure*}
    \centering
    \includegraphics[width=.9\linewidth]{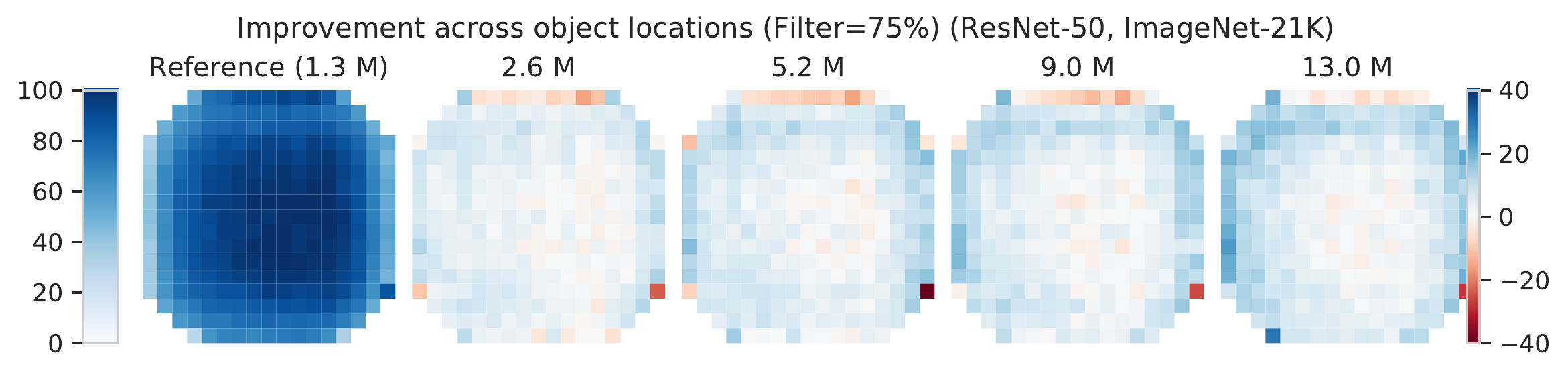}
    \includegraphics[width=.9\linewidth]{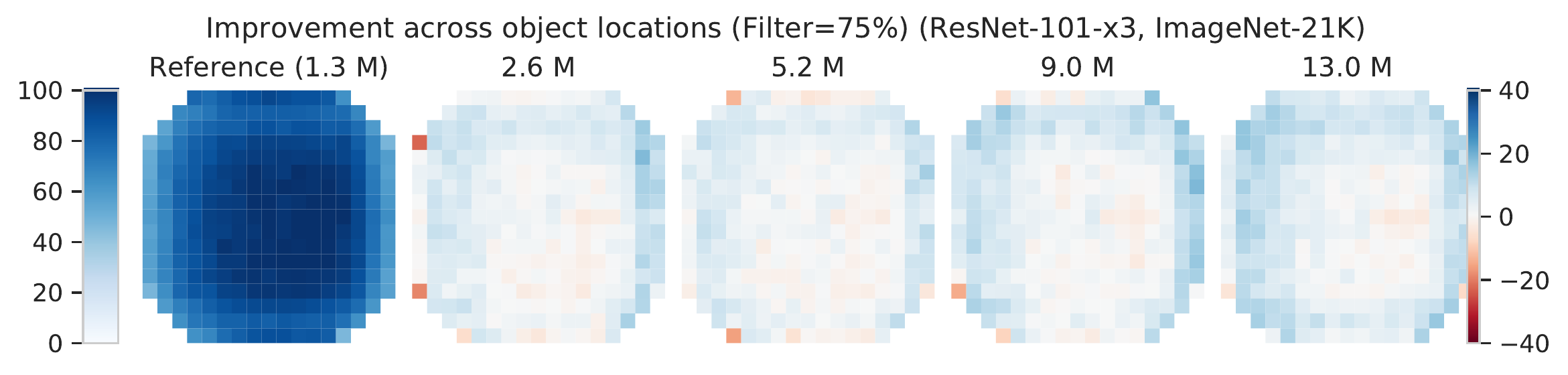}
    \includegraphics[width=.9\linewidth]{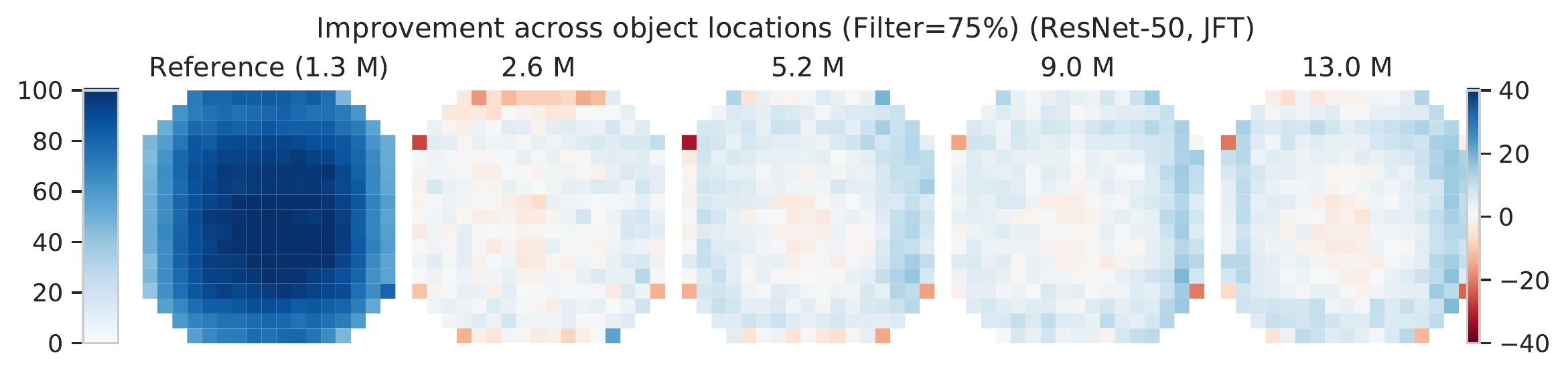}
    \includegraphics[width=.9\linewidth]{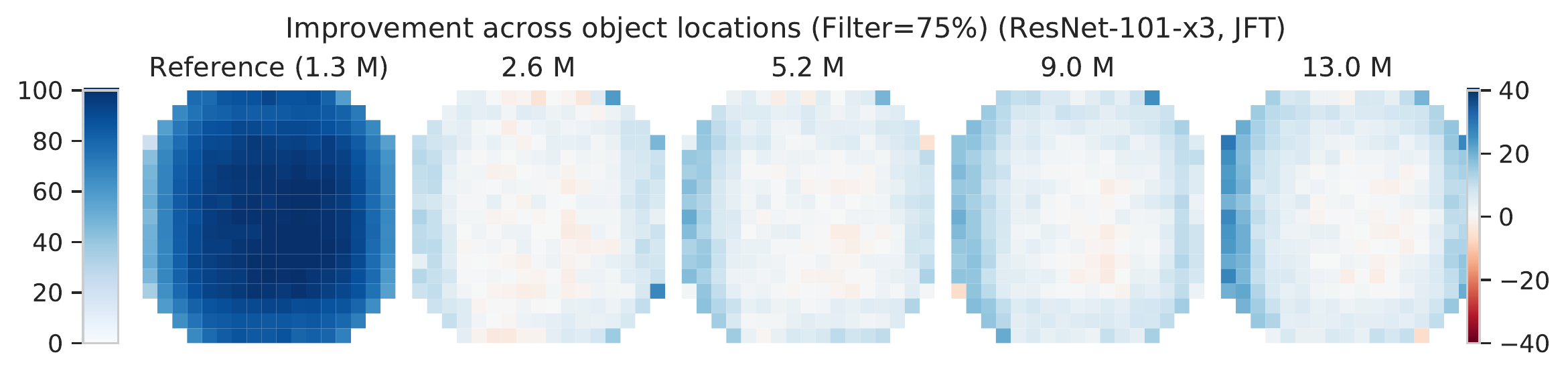}
    \caption{In the main paper, we presented results on the location dataset when not filtering out images where the objects were partially occluded, since that would exclude many locations from the dataset. For completeness, we present results filtering out objects that are less than 50\% or 75\% in the image in this figure and Figure~\ref{fig:synth-location-50pc}. \newline In the first column, for each location on the grid, we compute the average accuracy. Then, we normalize each location by the 95\textsuperscript{th} percentile across all locations, which quantifies the gap between the locations where the model performs well, and the ones where it under-performs (first column, dark blue vs white). Then, we consider models trained with more data, compute the same normalized score, and plot the \emph{difference} with respect to the first column. We observe that, as dataset size increases, sensitivity to object location decreases -- the outer regions improve in relative accuracy more than the inner ones (e.g. dark blue vs white on the second and third columns). The effect is harder to see since most pixels near the edges have been filtered out --- here we filter out all test images for which the foreground object is not at least $75\%$ within the image.
}
    \label{fig:synth-location-75pc}
\end{figure*}

\clearpage

\clearpage    
\section{Overview of model abbreviations}\label{app:model_overview}

\begin{table}[ht]
    \addtolength{\tabcolsep}{-3pt}  %
    \centering
    \resizebox{\textwidth}{!}{%
     \begin{tabular}{l l l r r r} 
     \toprule 
\textsc{Model name}	& \textsc{Type} & \textsc{Training data}	& \textsc{Architecture}	& \textsc{Depth}  & \textsc{Ch.} \\\midrule
\textsc{r50-imagenet-100}	 & \textsc{Supervised}	 & \textsc{ImageNet }	 & \textsc{ResNet}	 & \num{50}	 & \num{1}\\
\textsc{r50-imagenet-10}	 & \textsc{Supervised}	 & \textsc{ImageNet, 10\%}	 & \textsc{ResNet}	 & \num{50}	 & \num{1}\\
\textsc{bit-imagenet-r50-x1}	 & \textsc{Supervised \cite{bit}}	 & \textsc{ImageNet}	 & \textsc{ResNet}	 & \num{50}	 & \num{1}\\
\textsc{bit-imagenet-r50-x3}	 & \textsc{Supervised \cite{bit}}	 & \textsc{ImageNet}	 & \textsc{ResNet}	 & \num{50}	 & \num{3}\\
\textsc{bit-imagenet-r101-x1}	 & \textsc{Supervised \cite{bit}}	 & \textsc{ImageNet}	 & \textsc{ResNet}	 & \num{101}	 & \num{1}\\
\textsc{bit-imagenet-r101-x3}	 & \textsc{Supervised \cite{bit}}	 & \textsc{ImageNet}	 & \textsc{ResNet}	 & \num{101}	 & \num{3}\\
\textsc{bit-imagenet21k-r50-x1}	 & \textsc{Supervised \cite{bit}}	 & \textsc{ImageNet21k}	 & \textsc{ResNet}	 & \num{50}	 & \num{1}\\
\textsc{bit-imagenet21k-r50-x3}	 & \textsc{Supervised \cite{bit}}	 & \textsc{ImageNet21k}	 & \textsc{ResNet}	 & \num{50}	 & \num{3}\\
\textsc{bit-imagenet21k-r101-x1}	 & \textsc{Supervised \cite{bit}}	 & \textsc{ImageNet21k}	 & \textsc{ResNet}	 & \num{101}	 & \num{1}\\
\textsc{bit-imagenet21k-r101-x3}	 & \textsc{Supervised \cite{bit}}	 & \textsc{ImageNet21k}	 & \textsc{ResNet}	 & \num{101}	 & \num{3}\\
\textsc{bit-jft-r50-x1}	 & \textsc{Supervised \cite{bit}}	 & \textsc{JFT}	 & \textsc{ResNet}	 & \num{50}	 & \num{1}\\
\textsc{bit-jft-r50-x3}	 & \textsc{Supervised \cite{bit}}	 & \textsc{JFT}	 & \textsc{ResNet}	 & \num{50}	 & \num{3}\\
\textsc{bit-jft-r101-x1}	 & \textsc{Supervised \cite{bit}}	 & \textsc{JFT}	 & \textsc{ResNet}	 & \num{101}	 & \num{1}\\
\textsc{bit-jft-r101-x3}	 & \textsc{Supervised \cite{bit}}	 & \textsc{JFT}	 & \textsc{ResNet}	 & \num{101}	 & \num{3}\\
\textsc{bit-jft-r152-x4}	 & \textsc{Supervised \cite{bit}}	 & \textsc{JFT}	 & \textsc{ResNet}	 & \num{50}	 & \num{4}\\
\textsc{r50-imagenet-10-exemplar}	 & \textsc{Self-sup. \& cotraining \cite{s4l}}	 & \textsc{ImageNet, 10\%}	 & \textsc{ResNet}	 & \num{50}	 & \num{1}\\
\textsc{r50-imagenet-10-rotation}	 & \textsc{Self-sup. \& cotraining \cite{s4l}}	 & \textsc{ImageNet, 10\%}	 & \textsc{ResNet}	 & \num{50}	 & \num{1}\\
\textsc{r50-imagenet-100-exemplar}	 & \textsc{Self-sup. \& cotraining \cite{s4l}}	 & \textsc{ImageNet}	 & \textsc{ResNet}	 & \num{50}	 & \num{1}\\
\textsc{r50-imagenet-100-rotation}	 & \textsc{Self-sup. \& cotraining \cite{s4l}}	 & \textsc{ImageNet}	 & \textsc{ResNet}	 & \num{50}	 & \num{1}\\
\textsc{simclr-1x-self-supervised}	 & \textsc{Self-supervised \cite{simclr}, fine tuning}	 & \textsc{ImageNet}	 & \textsc{ResNet}	 & \num{50}	 & \num{1}\\
\textsc{simclr-2x-self-supervised}	 & \textsc{Self-supervised \cite{simclr}, fine tuning}	 & \textsc{ImageNet}	 & \textsc{ResNet}	 & \num{50}	 & \num{2}\\
\textsc{simclr-4x-self-supervised}	 & \textsc{Self-supervised \cite{simclr}, fine tuning}	 & \textsc{ImageNet}	 & \textsc{ResNet}	 & \num{50}	 & \num{4}\\
\textsc{simclr-1x-fine-tuned-10}	 & \textsc{Self-supervised \cite{simclr}, fine tuning}	 & \textsc{ImageNet, 10\%}	 & \textsc{ResNet}	 & \num{50}	 & \num{1}\\
\textsc{simclr-2x-fine-tuned-10}	 & \textsc{Self-supervised \cite{simclr}, fine tuning}	 & \textsc{ImageNet, 10\%}	 & \textsc{ResNet}	 & \num{50}	 & \num{2}\\
\textsc{simclr-4x-fine-tuned-10}	 & \textsc{Self-supervised \cite{simclr}, fine tuning}	 & \textsc{ImageNet, 10\%}	 & \textsc{ResNet}	 & \num{50}	 & \num{3}\\
\textsc{simclr-1x-fine-tuned-100}	 & \textsc{Self-supervised \cite{simclr}, fine tuning}	 & \textsc{ImageNet}	 & \textsc{ResNet}	 & \num{50}	 & \num{1}\\
\textsc{simclr-2x-fine-tuned-100}	 & \textsc{Self-supervised \cite{simclr}, fine tuning}	 & \textsc{ImageNet}	 & \textsc{ResNet}	 & \num{50}	 & \num{2}\\
\textsc{simclr-4x-fine-tuned-100}	 & \textsc{Self-supervised \cite{simclr}, fine tuning}	 & \textsc{ImageNet}	 & \textsc{ResNet}	 & \num{50}	 & \num{4}\\
\textsc{efficientnet-std-b0}	 & \textsc{Supervised \cite{tan2019efficientnet}}	 & \textsc{ImageNet}	 & \textsc{EfficientNet}	 & \num{18}	 & \num{1}\\
\textsc{efficientnet-std-b4}	 & \textsc{Supervised \cite{tan2019efficientnet}}	 & \textsc{ImageNet}	 & \textsc{EfficientNet}	 & \num{37}	 & \num{1}\\
\textsc{efficientnet-adv-prop-b0}	 & \textsc{Supervised \& adversarial \cite{xie2019adversarial}}	 & \textsc{ImageNet}	 & \textsc{EfficientNet}	 & \num{18}	 & \num{1}\\
\textsc{efficientnet-adv-prop-b4}	 & \textsc{Supervised \& adversarial \cite{xie2019adversarial}}	 & \textsc{ImageNet}	 & \textsc{EfficientNet}	 & \num{37}	 & \num{1}\\
\textsc{efficientnet-adv-prop-b7}	 & \textsc{Supervised \& adversarial \cite{xie2019adversarial}}	 & \textsc{ImageNet}	 & \textsc{EfficientNet}	 & \num{64}	 & \num{2}\\
\textsc{efficientnet-noisy-student-b0}	 & \textsc{Supervised \& distillation \cite{noisystudent}}	 & \textsc{ImageNet}	 & \textsc{EfficientNet}	 & \num{18}	 & \num{1}\\
\textsc{efficientnet-noisy-student-b4}	 & \textsc{Supervised \& distillation \cite{noisystudent}}	 & \textsc{ImageNet}	 & \textsc{EfficientNet}	 & \num{37}	 & \num{1}\\
\textsc{efficientnet-noisy-student-b7}	 & \textsc{Supervised \& distillation \cite{noisystudent}}	 & \textsc{ImageNet}	 & \textsc{EfficientNet}	 & \num{64}	 & \num{2}\\
\textsc{vivi-1x}	 & \textsc{Self-sup. \& cotraining \cite{vivi}}	 & \textsc{YT8M, ImageNet}	 & \textsc{ResNet}	 & \num{50}	 & \num{1}\\
\textsc{vivi-3x}	 & \textsc{Self-sup. \& cotraining \cite{vivi}}	 & \textsc{YT8M, ImageNet}	 & \textsc{ResNet}	 & \num{50}	 & \num{3}\\
\textsc{bigbigan-linear}	 & \textsc{Bidirectional adversarial \cite{bigbigan}}	 & \textsc{ImageNet}	 & \textsc{ResNet}	 & \num{50}	 & \num{1}\\
\textsc{bigbigan-finetune}	 & \textsc{Bidirectional adversarial \cite{bigbigan}}	 & \textsc{ImageNet}	 & \textsc{ResNet}	 & \num{50}	 & \num{1}\\
     \bottomrule
     \end{tabular}%
     }
    \caption{\label{tab:model_training_overview} Overview of models used in this study. \textsc{sup.} abbreviates for supervised pre-training. \textsc{Ch.} refers to the width multiplier for the number of channels.}
\end{table}

\end{document}